\documentclass[review,11pt]{elsarticle}
\usepackage{nomencl}
\makenomenclature
\usepackage{hyperref}
\usepackage{amssymb}
\usepackage{multirow}
\usepackage{amsmath} 
\usepackage{float}                  %
\usepackage{subfig}                 %
\usepackage{threeparttable}    

\biboptions{numbers,sort&compress}
\usepackage{geometry}
\usepackage{booktabs}
\usepackage{setspace}
\usepackage{makecell}
\usepackage{framed} 
\usepackage{amsmath}
\usepackage{longtable}
\usepackage{color}
\usepackage{cases}
\usepackage{enumitem}
\usepackage{booktabs} 
\journal{Advanced Engineering Informatics}

\begin{document}
	
	\begin{frontmatter}

		\title{Digital Twin-Driven Zero-Shot Fault Diagnosis of Axial Piston Pumps Using Fluid-Borne Noise Signals}
		
		\author[mymainaddress1,mymainaddress2]{Chang Dong}
		\author[mymainaddress1,mymainaddress2]{Jianfeng Tao\corref{mycorrespondingauthor}}
		\ead{chdong_sjtu@163.com}
		\author[mymainaddress1]{Chengliang Liu}
		\cortext[mycorrespondingauthor]{Corresponding author}
		\address[mymainaddress1]{State Key Laboratory of Mechanical System and Vibration, Shanghai Jiao Tong University, Shanghai 200240, China}
		\address[mymainaddress2]{Shanghai Platform for Smart Manufacturing Co.,Ltd, Shanghai 201306, China}

		\begin{abstract}
Axial piston pumps are crucial components in fluid power systems, where reliable fault diagnosis is essential for ensuring operational safety and efficiency. Traditional data-driven methods require extensive labeled fault data, which is often impractical to obtain, while model-based approaches suffer from parameter uncertainties. This paper proposes a digital twin (DT)-driven zero-shot fault diagnosis framework utilizing fluid-borne noise (FBN) signals. The framework calibrates a high-fidelity DT model using only healthy-state data, generates synthetic fault signals for training deep learning classifiers, and employs a physics-informed neural network (PINN) as a virtual sensor for flow ripple estimation. Gradient-weighted class activation mapping (Grad-CAM) is integrated to visualize the decision-making process of neural networks, revealing that large kernels matching the subsequence length in time-domain inputs and small kernels in time-frequency domain inputs enable higher diagnostic accuracy by focusing on physically meaningful features. Experimental validations demonstrate that training on signals from the calibrated DT model yields diagnostic accuracies exceeding 95\% on real-world benchmarks, while uncalibrated models result in significantly lower performance, highlighting the framework's effectiveness in data-scarce scenarios.
		\end{abstract}
		
	\end{frontmatter}
	
	
	\section{Introduction}
	\label{sec:introduction}
	
	Axial piston pumps serve as critical components in fluid power transmission systems and form the foundation for numerous high-stakes applications, such as aerospace, marine propulsion, and heavy machinery\cite{wang2025condition}. Their operational reliability directly determines the host system's overall safety and efficiency. These pumps convert mechanical energy into hydraulic power via the reciprocating motion of pistons within a rotating cylinder block. Yet, their performance and longevity face challenges from complex internal health dynamics, particularly across key friction pairs, such as the slipper/swashplate and piston/cylinder interfaces, which expose them to various faults\cite{dong2023subsequence}. Common failure modes—mechanical wear\cite{ma2025research}, cavitation damage\cite{ye2017experimental}, and increased leakage\cite{wang2023physics}—can trigger catastrophic system failures if left unaddressed. Thus, advanced condition monitoring and precise fault diagnosis prove essential for enabling predictive maintenance, reducing unscheduled downtime, and preserving operational integrity in hydraulic systems.

	Fault diagnosis approaches for axial piston pumps are generally classified into two primary categories\cite{dong2023subsequence,wang2025transfer}: model-based methods and data-driven methods. Various sensors are commonly employed for condition monitoring, including flow rate\cite{du2013layered}, temperature\cite{Ivantysyn2024ThermalIA}, high-frequency casing vibration\cite{xia2019spare}, high-frequency pressure ripple\cite{lu2015fault,lu2017fault}, and—more recently—dynamic cylinder block displacement \cite{xu2021direct,zhang2024cylinder}. Model-based methods are fundamentally grounded in physical modeling via single-physics or multi-physics simulations\cite{ye2025new} of the pump’s dynamics (encompassing multibody dynamics\cite{ye2021theoretical}, fluid dynamics\cite{dong2023inverse}, and elastohydrodynamic lubrication (EHL)\cite{lyu2020research,Ivantysyn2024ThermalIA}). The physical model in model-based methods naturally encompasses two interrelated directions. The first path focuses on degradation mechanism modeling, which examines long-term evolutionary processes within the pump. Given the vital role of the three primary friction pairs in sealing and lubrication, researchers have scrutinized their wear mechanisms most intensively. Notable examples include Lyu et al.'s \cite{lyu2020research} high-precision EHL-based prediction of piston/cylinder wear, validated experimentally with coordinate measuring machines to identify localized wear zones, and Long et al.'s \cite{long2024life} comprehensive life-prediction model for the slipper/swashplate pair. The second, more diagnostic path employs fault behavior modeling via fault injection and response analysis. Pioneering studies include Li et al.\cite{fs_Li2005piston}'s use of a lumped-parameter model, where they altered a single piston's diameter to simulate individual wear and replicated characteristic periodic pressure drops, and Bensaad et al.\cite{fs_bensaad2019new}'s observation of matching periodic pressure dips, paired with a Kalman filter to pinpoint the leaking piston. Nonetheless, early model-based approaches suffered from waveform mismatches between simulations and experiments, stemming from inadequate model fidelity and parameter uncertainties that hindered the practical application of model-based methods.

	 In recent years, data-driven methods have gained prominence by sidestepping modeling complexities: they treat the pump as a black box and derive mappings from monitored signals to health states\cite{lei2020applications}. Initial efforts depended on hand-engineered features—such as time- and frequency-domain statistics \cite{wang2018data} or EMD-based energy entropy \cite{lu2015fault,lu2017fault,wang2018data}—fed into shallow classifiers like support vector machines \cite{xia2019spare} and extreme learning machines \cite{lan2018fault}. The rise of deep learning has shifted focus to end-to-end frameworks, with 1D/2D CNNs \cite{tang2022novel,tang2021improved,tang2022intelligent,tang2022adaptive,tang2022synchrosqueezed,xia2025novel}, ResNets \cite{chen2024resnet,Liu2024TSattention}, attention-enhanced networks \cite{Liu2024TSattention,dong2025domain}, and similar architectures automatically extracting discriminative features from raw or spectrogram-transformed high-frequency signals. These approaches routinely exceed 90\% accuracy on laboratory fault datasets. For instance, Tang et al.\cite{tang2022intelligent} devised a normalized convolutional neural network with batch normalization to handle data distribution shifts, Bayesian optimization(BO) for hyperparameter tuning, and synchrosqueezed wavelet transform (SWT) preprocessing for vibration, pressure, and acoustic signals from a swash-plate axial piston pump. Their experiments showcased superior classification of five health states, surpassing traditional LeNet-5 and AlexNet architectures across signal types. Many studies have pursued novel network designs or preprocessing enhancements to boost accuracy. Extending single-fault detection, Liu et al.\cite{liu2024temporal} tackled compound faults via multi-sensor fusion, incorporating their innovation in cylinder block displacement signals, which provide superior fault sensitivity due to short propagation paths. To counter the class of exponential increase caused by compound fault, they employed binary classification strategies. Their temporal-spatial attention network, combined with this sensing approach, delivered peak accuracies in intricate fault scenarios.

	 Despite these advances, pure data-driven methods hinge on extensive labeled datasets from expensive and hazardous fault-injection experiments—a barrier for high-value applications such as aerospace or marine \cite{ivantysyn2018investigation}. As Ivantysyn et al.\cite{Ivantysyn2024ThermalIA} pointed out, the diversity of piston pump models and operating conditions demands fault-labeled datasets for every variant, rendering laboratory-tuned paradigms difficult to deploy in data-poor real-world settings. Moreover, the "black-box" nature of deep learning models complicates result interpretation\cite{jiang2025towards}. Building on established data-driven successes, transfer learning first emerged to lessen reliance on vast labeled data\cite{lei2020applications}. This technique transfers knowledge from a source domain to a data-scarce target domain. In axial piston pump diagnostics, two transfer paradigms dominate. The first bridges experimental datasets to address variations in operating conditions or pump models. Miao et al.\cite{miao2020application}, for example, sourced data from conventional hydraulic piston pumps and targeted seawater pump data, applying TrAdaBoost to secure over 90\% accuracy with just 30 target fault samples. He et al.\cite{he2022deep} similarly crafted an adversarial transfer framework fusing acoustic and vibration signals for robust diagnosis across nine operating conditions. The second paradigm is simulation-to-experiment transfer: synthetic data from simulation models enrich the source domain, while experimental data forms the target, fulfilling dual roles in augmenting scarce fault data and refining simulation-trained models to surmount historical model-based accuracy gaps. Liu et al.\cite{liu2025semi}, for instance, introduced a convolutional cross-branch architecture with self-attention for domain-invariant features, generating labeled source-domain pressure signals via AMESim lumped-parameter simulations of faults. Wang et al.\cite{wang2025transfer} advanced a transfer scheme using 3D CFD simulations for labeled faulty pressure signals as the source, followed by EMD preprocessing and a domain-adversarial neural network for experimental adaptation. Ma et al.\cite{ma2026multi} deployed an enhanced meta-transfer Learning network with MMD-driven adaptation; their co-simulation integrated multibody dynamics and hydraulics, injecting faults to yield vibration and pressure ripple signals, achieving over 98\% accuracy with only 10 labeled target samples and excelling in few-shot learning. Although simulation-to-experiment transfer eases fault dataset burdens, it still demands some experimental fault samples(i.e.,few-shot scenario) to reconcile simulated and real signals, limiting utility in ultra-data-scarce or zero-shot settings\cite{liu2023transfer}. While these methods often handle few-shot or unlabeled targets, they prove inadequate for high-stakes piston pump applications, where healthy-state data abounds but fault data does not.
	 
	 To address these constraints in severely data-limited scenarios, newer paradigms deem transfer learning inadequate and pivot to alternatives like digital twin (DT) frameworks in the face of zero-shot diagnostic scenarios. A DT creates a virtual replica of the physical asset, synchronized continuously with real-time data to mirror system dynamics\cite{tao2018digital}. For fault diagnosis, the DT refines physical model parameters—rather than black-box neural ones—using only healthy-state data, allowing diagnostic model training on simulated data alone. This yields precise diagnostics without experimental fault datasets and has demonstrated strong zero-shot performance against experimental faults in fields like bearing \cite{sobie2018simulation_bearing,gao2020fem_bearing,farhat2021digital_bearing}, gearbox \cite{xia2024digital_gear}, and hydraulic cylinder \cite{wang2022fault_cylinder} diagnostics. Xia et al.\cite{xia2024digital_gear}, for example, presented a DT-assisted gearbox framework employing variational mode decomposition and kurtosis-based features from simulated vibrations, bypassing measured fault data entirely. To our knowledge, DT-enabled prognostics and health management (PHM) for piston pumps in zero-shot contexts remains underexplored, with initial efforts focusing on thermal signals \cite{ivantysyn2018investigation,Ivantysyn2024ThermalIA,ivantysyn2024advancing}. Ivantysyn and Weber \cite{ivantysyn2018investigation,Ivantysyn2024ThermalIA,ivantysyn2024advancing} pioneered a high-fidelity lubrication-thermal DT via physics modeling, including thermal EHL gap models, implemented in Caspar FSTI software \cite{wieczorek2002computer,purdue_multics_2022}; they identified cylinder block surface temperature as a low-bandwidth efficiency indicator.
    
    Drawing on these previous zero-shot DT-based studies, this study proposes a novel DT-driven zero-shot fault diagnosis framework for axial piston pumps, leveraging FBN signals—which encompass high-frequency flow and pressure ripples\cite{opperwall2014combined}—as key indicators of outlet fluid dynamics. High-frequency pressure ripples prove easy to acquire, while flow ripples resist direct measurement with commercial meters. The framework begins by constructing a high-fidelity DT of the pump outlet fluid domain (typically pump-pipe-load). Calibration draws on our prior inverse transient analysis (ITA)-based pipe parameter method \cite{dong2023inverse}, validated for slipper and cylinder faults. Diagnostic models then train on simulated outlet flow ripples from a 3D CFD pump model and pressure ripples from a 1D unsteady flow model of pipelines, eschewing construction of physical fault dataset. Inputs for diagnostic models include raw time-domain signals or spectrograms. To estimate experimental outlet flow ripples, the framework applies our earlier PINN-based inverse solver\cite{dong2025innovative} within the calibrated pipeline model, enabling virtual sensing at the pump outlet. Finally, Grad-CAM verifies the diagnostic model's focus, aligning learned criteria with physical principles to inform neural network tuning. The principal contributions of this work are as follows:
\begin{enumerate}
	\item[(1)] We present an innovative DT-driven zero-shot fault diagnosis framework for axial piston pumps, which calibrates high-fidelity physical models using only healthy-state data and trains classifiers on simulated fault signals. This eliminates the need for expensive and risky experimental fault datasets while achieving diagnostic accuracies exceeding 95\% on experimental benchmarks, thereby addressing persistent challenges in data-limited, high-stakes applications.
	\item[(2)] We integrate Grad-CAM as a powerful tool to visualize the neural network's decision process and guide architecture design and parameter selection, enabling the preferential adoption of networks that consistently focus on physically meaningful fault signatures. Visualization results indicate that large kernels aligned with subsequence lengths in time-domain inputs and small kernels in time-frequency domain inputs yield superior accuracy, ensuring interpretability while directly informing architecture optimization to deliver highly robust and explainable diagnostic models that bolster confidence in PHM deployments for hydraulic systems.
\end{enumerate}
	
	The remainder of this paper is organized as follows. Section II establishes the theoretical foundations, covering data-driven fault diagnosis principles, Grad-CAM for interpretability, FBN physics, and the synchrosqueezed wavelet transform. Section III details the proposed digital twin-driven fault diagnosis framework. Section IV presents the experimental validation. Finally, Section V synthesizes the key findings, discusses practical implications, and outlines prospective research directions.
	
	\section{Theoretical background} 
	\subsection{Principles of experimental data-driven fault diagnosis}
	As a representative of data-driven approaches, deep learning excels in mechanical fault diagnosis by automatically extracting complex signal features, thereby overcoming the reliance on hand-engineered features in traditional methods~\cite{lei2020applications}. Fault diagnosis is often abstracted as a classification problem, aiming to map high-frequency monitoring signals to fault category labels. The formalized mathematical description is as follows:
	
	The fault diagnosis task is defined as learning a mapping function \( f: X \to Y \), where \( X \) denotes the input space, typically comprising spectrograms or raw signals, and \( Y \) represents the set of health state categories, e.g., \( Y = \{ \text{Healthy}, \text{Slipper Fault}, \text{Cylinder Fault}, \dots \} \). In deep learning-based methods, this function \( f \) is parameterized by a neural network, denoted as \( f_\theta \), where \( \theta \) are the network parameters. A dataset is constructed, and optimal parameters \( \theta^* \) are learned by minimizing the classification loss function:
	\begin{equation}
		L(\theta) = \sum_{(x, y) \in D} l(f_\theta(x), y),
		\label{eq:classification_loss}
	\end{equation}
	where \( l \) is typically the cross-entropy loss function, and the dataset consists of \( N \) samples: \( D = \{ (x^{(i)}, y^{(i)}) \}_{i=1}^{N} \), with \( y^{(i)} \) as the one-hot indicator variable for the \( i \)-th sample's true label in the \( k \)-th class. Upon training completion, inputting the signal to be evaluated into the neural network yields the diagnostic result via \( \hat{y} = \arg\max_k [f_\theta(x)]_k \), where \( [f_\theta(x)]_k \) denotes the \( k \)-th output logit or probability.
	
	Despite efficacy on labeled data, domain shifts (e.g., varying conditions) impair generalization. Transfer learning mitigates this by transferring knowledge from labeled source domain \( D_S = \{(x_S, y_S)\} \) to target domain \( D_T \), which includes limited labeled data \( D_T^l = \{(x_T^l, y_T^l)\} \) (possibly empty) and unlabeled data \( D_T^u = \{x_T^u\} \) for domain alignment. A pre-trained \( f_{\theta_S} \) initializes \( \theta_S \), fine-tuned via\cite{lei2020applications}:
	\begin{equation}
		L_{TL}(\theta) = \sum_{(x_S, y_S) \in D_S} l(f_\theta(x_S), y_S) + \lambda \sum_{(x_T^l, y_T^l) \in D_T^l} l(f_\theta(x_T^l), y_T^l) + \mu L_{DA}(g_\theta(D_S), g_\theta(D_T^u)),
		\label{eq:transfer_loss}
	\end{equation}
	where \( \lambda, \mu > 0 \) weight target supervision and domain adaptation loss \( L_{DA} \) (e.g., adversarial divergence or maximum mean discrepancy via a domain discriminator \( g_\theta \)), aligning distributions of source features \( g_\theta(D_S) = \{g_\theta(x_S)\} \) and unlabeled target features \( g_\theta(D_T^u) = \{g_\theta(x_T^u)\} \). This promotes domain-invariant representations, facilitating adaptation with few-shot labeled targets and unlabeled target data in diagnostics.

	\subsection{Grad-CAM for Model Interpretability}
	Grad-CAM\cite{selvaraju2017grad} advances interpretable deep learning by producing heatmaps of discriminative input regions, extending Class Activation Mapping (CAM) for versatile architectures without alterations. It employs gradients of the target class score \( y_c \)—the \( c \)-th element of \( f_\theta(x) \) (refined via Eq.~\eqref{eq:classification_loss} or Eq.~\eqref{eq:transfer_loss})—relative to final convolutional maps, illuminating decisions. Weights are calculated by spatial gradient averaging\cite{selvaraju2017grad}:
	\begin{equation}
		\alpha^c_k = \frac{1}{Z} \sum_i \sum_j \frac{\partial y_c}{\partial A_k^{ij}},
		\label{eq:gradcam_weights}
	\end{equation}
	yielding a localization map\cite{selvaraju2017grad}:
	\begin{equation}
		L^c_{\text{Grad-CAM}} = \text{ReLU} \left( \sum_k \alpha^c_k A_k \right),
		\label{eq:gradcam_map}
	\end{equation}
	with ReLU emphasizing positive influences; the map overlays inputs post-upsampling.
	\subsection{Fluid-borne noise signal}
	FBN encompasses pressure and flow ripples that propagate through hydraulic fluid in power systems, arising primarily from the intermittent displacement action of positive displacement units, such as axial piston pumps and external gear machines\cite{opperwall2014combined}. These ripples stem from kinematic effects (due to finite displacement chambers), compressible backflow during port transitions, and leakage across friction interfaces\cite{opperwall2014combined,dong2023inverse}. Faults—particularly leakage failures, which create new flow paths—exacerbate FBN by inducing short-duration leaks, resulting in abrupt, abnormal flow reductions that distort pulsation waveforms\cite{yang2008condition,dong2025innovative}. Upon interacting with pipeline impedance, these ripples are amplified or attenuated through transient wave propagation, complicating direct health assessment. For instance, Yang et al.\cite{yang2008condition} demonstrated measurable flow ripple anomalies in vane pumps under cavitation-induced side-plate erosion using ISO-standard measurement techniques, underscoring FBN's sensitivity to early degradation of pumps. The propagation of these transient waves is governed by the one-dimensional unsteady flow equations (assuming \( Q/A_L \ll a \))\cite{dong2023inverse}:
	\begin{equation}
		\frac{\partial p}{\partial x} + \frac{\rho}{A} \frac{\partial Q}{\partial t} + f(Q) = 0,
		\label{eq:momentum}
	\end{equation}
	\begin{equation}
		\frac{\partial p}{\partial t} + \frac{\rho a^2}{A} \frac{\partial Q}{\partial x} = 0,
		\label{eq:continuity}
	\end{equation}
	where \( p(x,t) \) and \( Q(x,t) \) denote transient pressure and flow rate, respectively; \( a \) is the wave speed; \( x \) the axial coordinate; \( \rho \) the fluid density; and \( A = \pi D^2/4 \) the pipe cross-sectional area (with diameter \( D \)). The friction term \( f(Q) \) is commonly modeled as steady laminar loss:
	\begin{equation}
		f(Q) = \frac{128 \rho \nu_0}{\pi D^4} Q,
	\end{equation}
	where \( \nu_0 \) is the dynamic viscosity.
	\subsection{Synchrosqueezed Wavelet Transform}
	The SST\cite{daubechies2011synchrosqueezed} refines the continuous wavelet transform to achieve superior time-frequency resolution for non-stationary signals, such as hydraulic pressure ripples, by reassigning wavelet coefficients to their instantaneous frequencies, thereby sharpening energy ridges and mitigating CWT's frequency smearing at low scales.
	
	For a signal \( s(t) \), the CWT is defined as \( W_s(a,b) = \int s(t) \psi^*\left(\frac{t-b}{a}\right) dt \), where \( \psi \) is the mother wavelet, \( a > 0 \) the scale, and \( b \) the time shift. SWT reallocates via the instantaneous frequency estimate \( \omega_s(a,b) = \Re\left(\frac{\partial_t W_s(a,b)}{2\pi i W_s(a,b)}\right) \) (for \( |W_s(a,b)| > \epsilon \)), yielding the synchrosqueezed representation\cite{daubechies2011synchrosqueezed}:
	\begin{equation}
		T_s(a,b) = \iint_{a'} W_s(a',b) \delta(\omega_s(a',b) - 1/a) \frac{da'}{a'^2},
	\end{equation}
	where \( \delta \) is the Dirac delta function, concentrating energy along true frequency paths. 

		\section{Proposed Method}
		\subsection{Motivation for Digital Twin-Driven Diagnosis}
		Although data-driven fault diagnosis has achieved significant progress, its practical deployment is often hindered by the substantial cost and risk involved in acquiring comprehensive labeled fault datasets. This challenge is particularly pronounced for critical high-value assets such as axial piston pumps, for which collecting run-to-failure data is prohibitively expensive and frequently hazardous. While transfer learning offers a means to mitigate data scarcity, it still relies on the availability of some fault data from the target domain—whether few-shot labeled ($D_T^l$) or unlabeled ($D_T^u$)—to achieve effective domain adaptation, as formalized in Eq.~\eqref{eq:transfer_loss}. Moreover, as noted earlier, the pressure ripple signals of a piston pump are influenced not only by the health state of the pump but also by the impedance characteristics of the connected pipeline system~\cite{dong2023inverse}. Consequently, in zero-shot scenarios where experimental fault samples are entirely unavailable, these requirements remain a major barrier.
		
		To address this fundamental limitation, we propose a DT–driven zero-shot fault diagnosis paradigm. The core idea is to employ a high-fidelity physics-based simulation model $\mathcal{M}$ as a generative surrogate for experimental fault data. Crucially, the model parameters $\mathbf{\Theta}$ are calibrated using only healthy-state measurements, which are typically readily available, thereby ensuring the digital twin accurately captures the dynamic behavior of the target system. This calibration process minimizes the discrepancy between the simulated and measured healthy-state signals:
		\begin{equation}
			\mathbf{\Theta}^* = \arg\min_{\mathbf{\Theta}} \mathcal{L}_{\text{cal}}\left( x_{\text{simu}}^{H}(\mathbf{\Theta}),\ x^{H}_{\text{meas}} \right)
			\label{eq:model_calibration}
		\end{equation}
		where $\mathcal{L}_{\text{cal}}$ is a suitable loss function (e.g., $L^2$ norm). Once calibrated, the $\mathcal{M}(\cdot; \mathbf{\Theta}^*)$ can simulate the system's response for any predefined fault condition $y$, generating a synthetic dataset:
		\begin{equation}
			D_{\text{simu}} = \left\{ \left( \mathcal{M}(y^{(i)}; \mathbf{\Theta}^*),\ y^{(i)} \right) \right\}_{i=1}^{K}
			\label{eq:synthetic_dataset}
		\end{equation}
		where
		\begin{equation}
			x^y_{\text{simu}} = \mathcal{M}(y, \mathbf{\Theta}^*),
		\end{equation}
	where $K$ denotes the number of considered fault states. This approach establishes a principled methodology for learning from physical principles, effectively transforming the zero-shot diagnosis problem into a task of model calibration and simulation. It can be viewed as calibrating a physics-based model using only healthy labeled data from the target domain to extract domain-invariant features—a concept analogous to domain adaptation in transfer learning. However, unlike purely data-driven methods that learn such features through implicit neural mappings and often require few-shot fault data or unlabeled fault data for distribution alignment, the invariance in our framework is explicitly anchored and governed by the underlying physical model.
		
		\subsection{Overview of The Proposed Diagnostic Framework}
The proposed framework, illustrated in Fig.~\ref{fig_flow_chart1}, consists of an offline DT construction/calibration phase and an online diagnostic phase. The detailed workflow is as follows:

		\begin{figure}[!h]
			\centering
			\includegraphics[width=1\columnwidth]{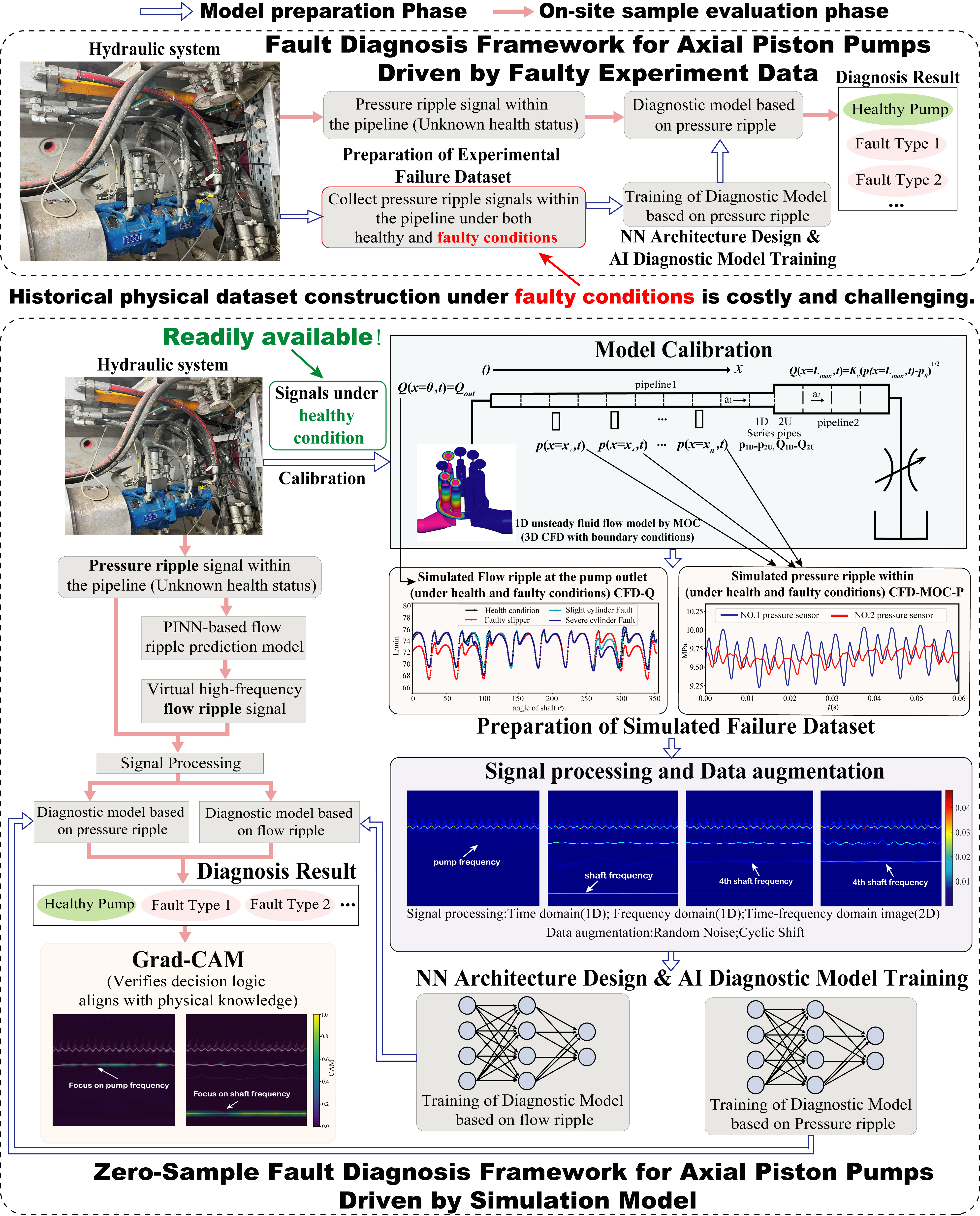}
			\caption{The proposed DT-driven zero-shot fault diagnosis framework for axial piston pumps using FBN signal}
			\label{fig_flow_chart1}
		\end{figure}

	\begin{enumerate}[label=\textbf{Step \arabic*}:, leftmargin=*, itemsep=5pt]
		\item \textbf{Healthy-State Data Acquisition}: Acquire pressure ripple signals $p_{\text{meas}}^H(t)$ from the target hydraulic system under known healthy operating conditions.
		\item \textbf{Digital Twin Construction \& Calibration}: Construct simulation models $\mathcal{M} = \{\mathcal{M}_{\text{3D-CFD}},\\ \mathcal{M}_{\text{1D-MOC}}\}$ representing the dynamics of FBN. The high-fidelity 3D CFD model $\mathcal{M}_{\text{3D-CFD}}$ simulates the pump's internal flow to provide the pump outlet flow ripple $Q_{\text{out}}^H(t)$. This serves as the boundary condition for the efficient 1D unsteady pipeline model $\mathcal{M}_{\text{1D-MOC}}$. Calibrate the model parameters via ITA by solving the optimization problem in Eq.~\eqref{eq:model_calibration}, minimizing the error between $\mathcal{M}_{\text{1D-MOC}}(Q_{\text{out}}^H; \mathbf{\Theta})$ and $p_{\text{meas}}^H(x_s,t)$ at the sensor location $x_s$.
		\item \textbf{Synthetic Fault Data Generation}: Inject synthetic fault into $\mathcal{M}_{\text{3D-CFD}}$ to simulate pump outlet flow ripples $Q_{\text{out}}^y(t)$ for each fault type $y$. Feed $Q_{\text{out}}^y(t)$ into the calibrated $\mathcal{M}_{\text{1D-MOC}}(\cdot; \mathbf{\Theta}^*)$ to generate corresponding pressure ripple signals $p^y(x_s, t)$. This yields paired synthetic samples $(x_{\text{simu}}^y, y)$, where $x_{\text{simu}}^y$ can be either the flow ripple $Q_{\text{out}}^y(t)$ or the pressure ripple $p^y(x_s, t)$.
		\item \textbf{Database Construction \& Preprocessing}: To enhance data diversity and model robustness, apply data augmentation to the periodic simulation signals. For a simulated signal $x_{\text{simu}} = [x_1, x_2, \dots, x_T]$ representing one pump cycle, generate augmented samples via cyclic shift and additive noise. Preprocess all signals by mean removal and standardization. For models requiring time-frequency inputs, apply the SWT to convert $x_{\text{simu}}^y$ into a spectrogram $S^y(\tau, \omega)$. The final simulation database is $D_{\text{simu}} = \{(S^y \text{ or } x_{\text{simu}}^y,\ y)\}$.
		\item \textbf{Diagnostic Model Training}: Design and train a deep learning classifier $f_\theta$ (e.g., 1D-CNN for time-series or 2D-CNN for spectrograms) using exclusively the preprocessed simulation database $D_{\text{simu}}$.
		\item \textbf{Online Evaluation with Experimental Data}:
		\begin{itemize}
			\item \textit{Pressure Ripple-Based Diagnosis}: Directly preprocess the experimental pressure signal $p_{\text{exp}}(t)$ (applying the same mean removal, scaling, and SWT as in Step 4) and input it into the trained model $f_{\theta^*}$ for classification.
			\item \textit{Flow Ripple-Based Diagnosis}: First, employ a PINN-based inverse solver $\mathcal{G}_{\text{PINN}}$ to estimate the pump outlet flow ripple from the experimental pressure signal $p_{\text{exp}}(x_s,t)$: $\hat{Q}_{\text{out}}(t) = \mathcal{G}_{\text{PINN}}(p_{\text{exp}}(x_s,t); \mathbf{\Theta}^*)$. Then, preprocess $\hat{Q}_{\text{out}}(t)$ and input it into the corresponding flow ripple-trained diagnostic model $f_{\theta^*}$.
		\end{itemize}
		\item \textbf{Interpretability Analysis with Grad-CAM}: Apply Grad-CAM to the trained model $f_{\theta^*}$ to generate visual heatmaps $L^c_{\text{Grad-CAM}}$ (Eq.~\eqref{eq:gradcam_map}) for a given prediction of class $c$. This highlights the regions in the input signal (time‑domain or time‑frequency) most influential for the decision, thereby determining whether the network has learned physically meaningful and domain‑invariant features. The resulting visualization thus serves as a direct guide for selecting and optimizing neural network architectures (e.g., kernel size, depth), enabling engineers to prioritize the design of models that reliably attend to genuine fault signatures over spurious correlations.

	\end{enumerate}

	\subsection{Detailed Implementation of Key Steps}
	We now proceed to describe the approach in detail. High-frequency pressure ripple signals are acquired under healthy operating conditions, as direct measurement of flow ripple remains challenging. To capture the full wave propagation dynamics, pressure sensors are positioned at multiple locations along the pipeline wherever feasible. A physical model is then built to describe FBN, with the common structure generally being pump-pipe-load; the load may vary, such as a hydraulic valve, hydraulic cylinder, or hydraulic motor. The physical system studied in this work is a pump-pipe-load configuration; in this study, a pump-pipe-valve setup is adopted, consistent with our prior work~\cite{dong2023inverse}. As shown in our previous research\cite{dong2023inverse}, we first used a 3D CFD model to describe the FBN signal. This model covers the pump-pipe-valve system and uses the orifice equation as the boundary condition to simulate the valve as the load. To balance computational efficiency with physical fidelity in previous work, we proposed using the flow ripple at the pump outlet $Q_{\text{out}}^H(t)$ obtained from the 3D CFD solution $\mathcal{M}_{\text{3D-CFD}}$ as the boundary condition for the one-dimensional unsteady flow model $\mathcal{M}_{\text{1D-MOC}}$ within pipeline system, thus accelerating the pipeline parameter calibration process. The pipeline is discretized into $M$ segments, each defined by wave speed $a_i$, diameter $D_i$, and length $L_i$. With $N$ pressure sensors at positions $x_j$ and a load characterized by the valve coefficient $K_v$, the parameter vector is $\mathbf{\Theta} = \{a_i, D_i, L_i, x_j, K_v\}$. Calibration is performed via ITA, formulated as the multi-objective optimization\cite{dong2023inverse}:
	\begin{equation}
		\mathbf{\Theta}^* = \arg \min_{\mathbf{\Theta}} f(\mathbf{\Theta}), \quad \text{where} \quad f_s(\mathbf{\Theta}) = |p_{\text{meas}}^H(x_s,t) - p^H(x_s, t)|_2^2, \ s=1,\dots,N.
		\label{eq:ita_optimization}
	\end{equation}
	where $p_{\text{meas}}^H(x_s,t)$ and $p^H(x_s, t)$ is the measured and simulated pressure signal under health pump state, respectively. Table \ref{table_calibrated_model} presents the optimized parameters obtained for the experimental setup used in previous study. Further details are available in~\cite{dong2023inverse}.

	With the calibrated parameters $\mathbf{\Theta}^*$ fixed, synthetic fault data are generated by virtually injecting faults into $\mathcal{M}_{\text{3D-CFD}}$. For each fault type $y$ (e.g., slipper wear or cylinder leakage) and severity level, the fluid domain geometry is modified to introduce the corresponding leakage paths. The resulting pump outlet flow ripple $Q_{\text{out}}^y(t)$ is then propagated through the calibrated $\mathcal{M}_{\text{1D-MOC}}(\cdot;\mathbf{\Theta}^*)$ to obtain the corresponding pressure ripple $p^y(x_s,t)$ at the sensor locations. This yields two labeled synthetic datasets: flow-ripple-based $\{(Q_{\text{out}}^y(t), y)\}$ and pressure-ripple-based $\{(p^y(x_s,t), y)\}$.
	
	The raw simulated signals are noise-free and represent a single pump rotation cycle. To create a diverse and robust training set $D_{\text{simu}}$, data augmentation is applied. Exploiting the periodicity of the signals under steady-state operation, cyclic shifts are performed: for a signal $x_{\text{simu}} = [x_1, \dots, x_T]$, a $k$-step shift produces $[x_{k+1}, \dots, x_T, x_1, \dots, x_k]$. Additionally, white Gaussian noise is added to emulate measurement noise and minor operational variations.  Related research\cite{gao2020fem_bearing,liu2023transfer} uses generative adversarial networks for data augmentation of simulation data, but since this paper has already achieved high accuracy through cyclic shifting and random noise methods, GAN-based method was not used. For time-frequency-based diagnostic models, the augmented signals are transformed using the SWT with a Morse mother wavelet, tuned to resolve the characteristic shaft and piston harmonics.

A unified data preprocessing pipeline is implemented for both simulated and experimental signals to ensure robust and generalizable diagnostic model performance while preserving physically discriminative fault signatures. Z-score normalization is avoided, as it would eliminate inter-class amplitude differences—a critical diagnostic indicator, since leakage faults reduce ripple magnitude due to increased internal bypass flow. For time-domain signals, preprocessing involves only mean removal for each ripple signal \(x(t)\), computed as \(\mu_x = \frac{1}{T} \sum_{t=1}^T x(t)\), yielding \(\bar{x}(t) = x(t) - \mu_x\). This is applied consistently to both synthetic training data and experimental test data, maintaining relative amplitude differences across health states. For time-frequency domain inputs, the augmented signals are first transformed using the SST with a Morse mother wavelet to generate spectrograms \(S^y(\tau, \omega)\), tuned to resolve characteristic shaft and piston harmonics. To focus on the primary energy distribution, the frequency range is limited to below ten times the pumping frequency (i.e., 1500 Hz at 1000 r/min and 9 pistons)\cite{dong2023inverse}. The spectrogram magnitudes are then multiplied by a scaling coefficient to compress intensity values and prevent saturation in neural network inputs. Following this, an exponential transformation is applied to the scaled magnitudes (i.e., \(\exp(k \cdot S^y(\tau, \omega))\). A logarithmic transformation is subsequently applied to the frequency axis to evenly distribute key frequency points of interest (multiples of shaft and pumping frequencies) across the spectrogram. Finally, the transformed spectrogram is resized to a 256×256 image via linear interpolation to serve as input for 2D CNN models..

For flow-ripple-based diagnosis, the experimental pump outlet flow ripple is first estimated using the PINN inverse solver $\mathcal{G}_{\text{PINN}}$ developed in our prior work~\cite{dong2025innovative}. The PINN approximates the pressure and flow fields $u_\theta(x,t) = [p_\theta(x,t), Q_\theta(x,t)]$ and is trained by minimizing
\begin{equation}
	\mathcal{L}(\theta) = \mathcal{L}_{\text{data}}(\theta) + \mathcal{L}_{\text{pde}}(\theta) + \mathcal{L}_{\text{bc}}(\theta),
\end{equation}
with physical parameters fixed to $\mathbf{\Theta}^*$. The data loss $\mathcal{L}_{\text{data}}$ enforces agreement with measured pressure $p_{\text{exp}}(x_s,t)$ at sensor locations, while $\mathcal{L}_{\text{pde}}$ and $\mathcal{L}_{\text{bc}}$ enforce the governing equations and known boundary conditions. The estimated flow ripple $\hat{Q}_{\text{out}}(t) = Q_\theta(0,t)$ is then subjected to the same preprocessing steps (mean removal, reference scaling, and optional SWT) before classification. In both diagnostic pathways (pressure-ripple or estimated flow-ripple), the preprocessed experimental signal is fed to the classifier trained exclusively on $D_{\text{simu}}$, yielding the predicted health state $\hat{y} = \arg\max_k [f_{\theta^*}(\bar{x})]_k$.
	
 Finally, to ensure the diagnostic model does not rely on spurious correlations, Grad-CAM is employed. By visualizing the resulting activation maps on experimental inputs, we examine whether the network consistently attends to physically interpretable features—such as characteristic dips in flow ripple or distinct harmonic distortions associated with leakage faults. This analysis validates that the model's decision-making aligns with established physical principles, thereby confirming the consistency and reliability of the neural network architecture.
 
	\section{Results and discussion}
	
	\subsection{Experimental setup and Dataset construction}

\begin{figure}[!h]
		\centering	
		\includegraphics[width=0.85\columnwidth]{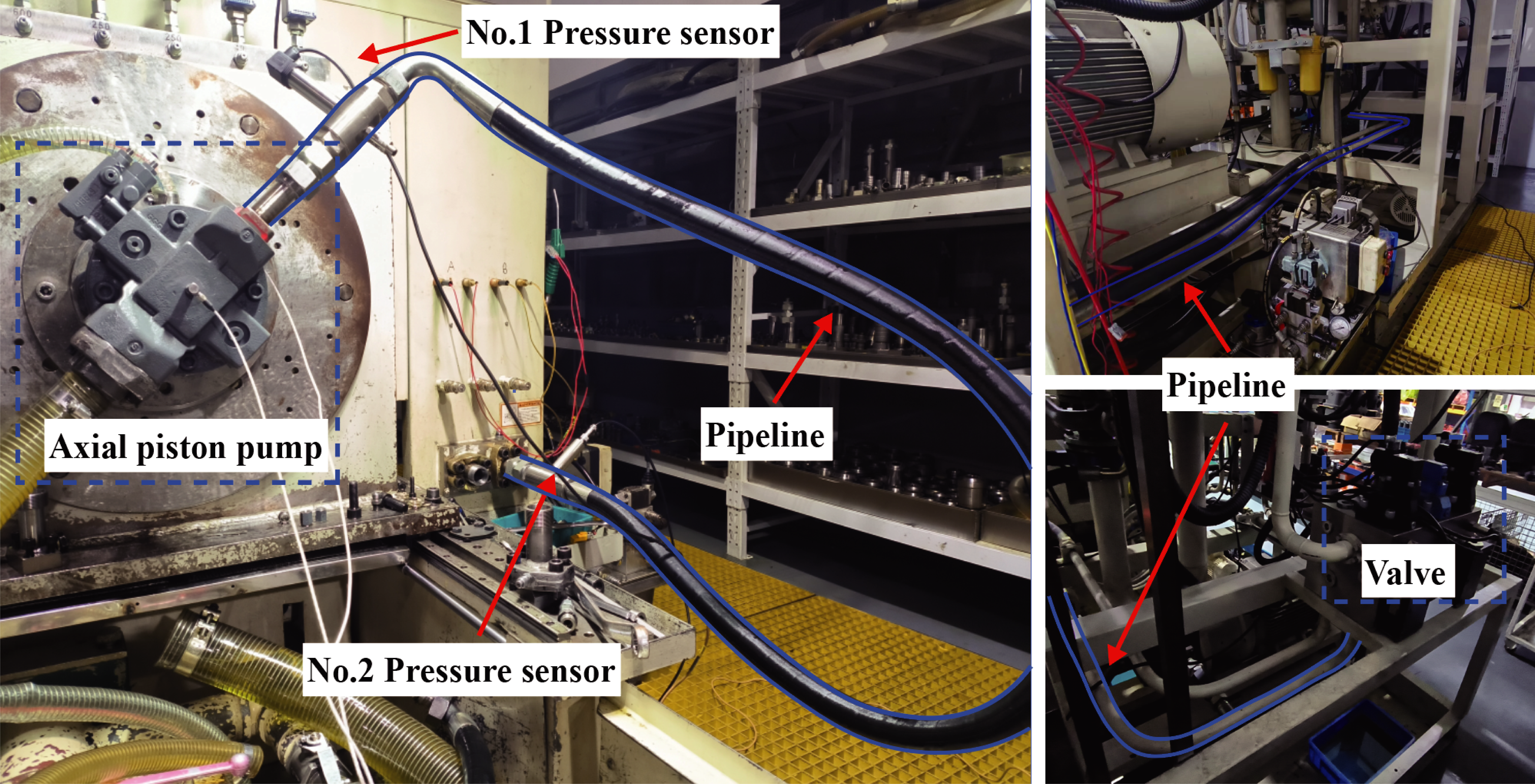}
		\caption{Test Rig}
		\label{fig_test_rig}
	\end{figure}

	The experimental setup adopted in this study, as illustrated in Fig.\ref{fig_test_rig}, is consistent with that employed in our previous investigations \cite{dong2023inverse,dong2025innovative}. As outlined in Section 3, the initial step entails acquiring pressure ripple data under healthy operating conditions. Two high-frequency pressure sensors, each with a bandwidth of 20 kHz, are positioned at the upstream and midstream sections of the pipeline. Data acquisition is performed using the DH5902N device at a sampling rate of 51,200 Hz. For further details regarding the experimental configuration, readers are directed to \cite{dong2023inverse}.

	 \begin{figure}[!h]
		\centering	
		\includegraphics[width=1\columnwidth]{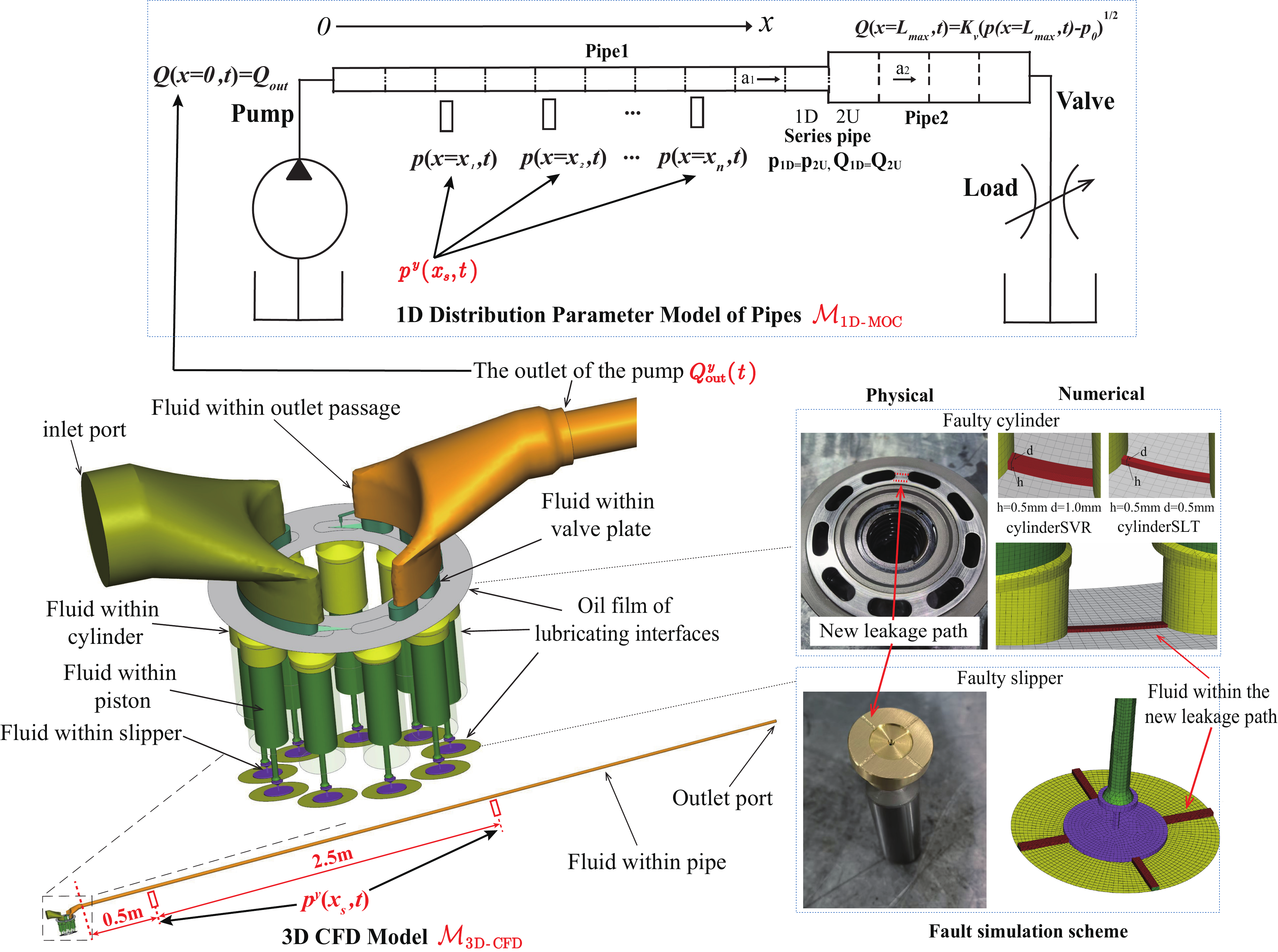}
		\caption{Numerical model and Fault injection scheme}
		\label{fig_simulation_model}
	 \end{figure}

	The subsequent phase involves constructing and calibrating a digital twin model that accurately replicates the system's behavior under healthy conditions. This model was developed in our prior work; comprehensive details are available in \cite{dong2023inverse}. As depicted in Fig.\ref{fig_simulation_model}, a 3D CFD model incorporating pump-pipe(5m)-valve(orifice equation as boundary condition) serves as $\mathcal{M}_{\text{3D-CFD}}$. The pump outlet flow ripple $Q_{\text{out}}^y(t)$ derived from this model is subsequently applied as the boundary condition for the one-dimensional unsteady flow model $\mathcal{M}_{\text{1D-MOC}}$, yielding the pressure ripple $p^y(x_s,t)$ at sensor locations. Calibration of $\mathcal{M}_{\text{1D-MOC}}$ is achieved using the pump outlet flow ripple $Q_{\text{out}}^H(t)$ under healthy conditions (i.e., without fault injection), employing ITA. The calibrated parameters are presented in Table \ref{table_calibrated_model}. Fig. \ref{fig_simulation_model}(b) compares the simulated signals from the calibrated $\mathcal{M}_{\text{1D-MOC}}$ with measurements obtained from the experimental test rig (as shown in Fig. \ref{fig_test_rig}). Fig. \ref{fig_simulation_model}(a) contrasts the simulated pressure ripples from $\mathcal{M}_{\text{3D-CFD}}$ at 0.5 m and 3 m with the corresponding experimental data, representing results prior to parameter calibration. Notably, the calibrated $\mathcal{M}_{\text{1D-MOC}}$ exhibits excellent agreement with the measured signals. In contrast, the uncalibrated 3D CFD outputs display discrepancies in amplitude and phase, which could compromise diagnostic accuracy if used for diagnostic purposes.

\begin{figure}[!htbp]
	\centering	
	\includegraphics[width=0.95\columnwidth]{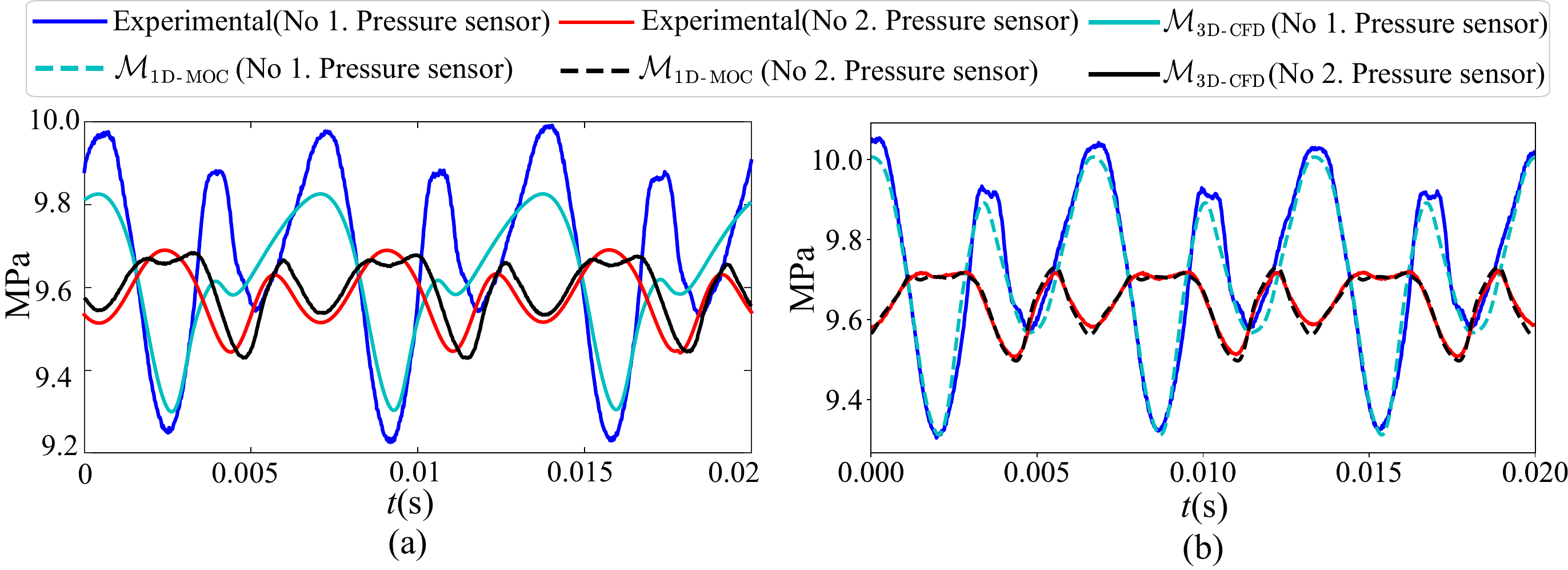}
	\caption{Pressure ripple comparison under health conditions: experimental data versus simulation results from (a) $\mathcal{M}_{\text{3D-CFD}}$(without ITA calibration) and (b) $\mathcal{M}_{\text{1D-MOC}}$(with ITA calibration).}
	\label{fig_pressure_ripple_H}
  \end{figure}

	The second step involved in the method mentioned is to establish and calibrate a digital twin model that matches the healthy state well. This has been established in our previous work. For more details, please refer to reference\cite{dong2023inverse}. As shown in the figure, a 3D CFD model with boundary conditions of pump-5m pipe-valve is used as $\mathcal{M}_{\text{3D-CFD}}$ in the lower left corner. Then, $Q_{\text{out}}^y(t)$ obtained at the pump outlet is used as the boundary condition of the 1D unsteady model $\mathcal{M}_{\text{1D-MOC}}$, and $p^y(x_s,t)$ at each sensor location is input. Then, the flow signal $Q_{\text{out}}^H(t)$ obtained at the pump outlet under the condition of no injection fault (i.e., pump health state) is used, and then $p^H(x_s,t)$ is obtained from $\mathcal{M}_{\text{1D-MOC}}$ to calibrate the parameters $\mathbf{\Theta}^*$ of $\mathcal{M}_{\text{1D-MOC}}$. The table shows the calibrated parameters based on ITA. Fig.\ref{fig_pressure_ripple_H} (b) shows a comparison between the simulated signal output by $\mathcal{M}_{\text{1D-MOC}}$ after parameter calibration and the measured signal on the experimental platform (as shown in the figure). Fig.\ref{fig_pressure_ripple_H} (a) shows a comparison between the simulated pressure ripple signals at 0.5m and 3m of $\mathcal{M}_{\text{3D-CFD}}$ and the measured signals, representing the results without parameter calibration. It can be seen that the simulated pressure ripple signal output by the calibrated $\mathcal{M}_{\text{1D-MOC}}$ matches the measured signal well. However, the signal output directly from the 3D CFD model without parameter calibration has differences in amplitude and phase, which may lead to poor accuracy when training based on this signal.
	\begin{table}[!h]
		\caption{Calibrated parameters $\mathbf{\Theta}^*$  for the model}
		\centering
		\renewcommand{\arraystretch}{1.2}
		\begin{tabular}{ll}
			\toprule
			Parameters & Calibrated Value \\
			\midrule
			Wave speeds ($a_1, a_2, a_3$) & 1022.9 m/s, 982.7 m/s, 1308.4 m/s \\
			Diameters ($D_1, D_2, D_3$) & 19.71 mm, 23.07 mm, 14.86 mm \\
			Lengths ($L_1, L_2, L_3$) & 5.953 m, 1.144 m, 1.675 m \\
			Sensor positions ($x_1, x_2$) & 0.4167 m, 2.5895 m \\
			Load Pressure ($P_{Set}$) & 9.484 MPa \\
			\bottomrule
		\end{tabular}
		\label{table_calibrated_model}
	\end{table}

\begin{figure}[!h]
		\centering	
		\includegraphics[width=0.75\columnwidth]{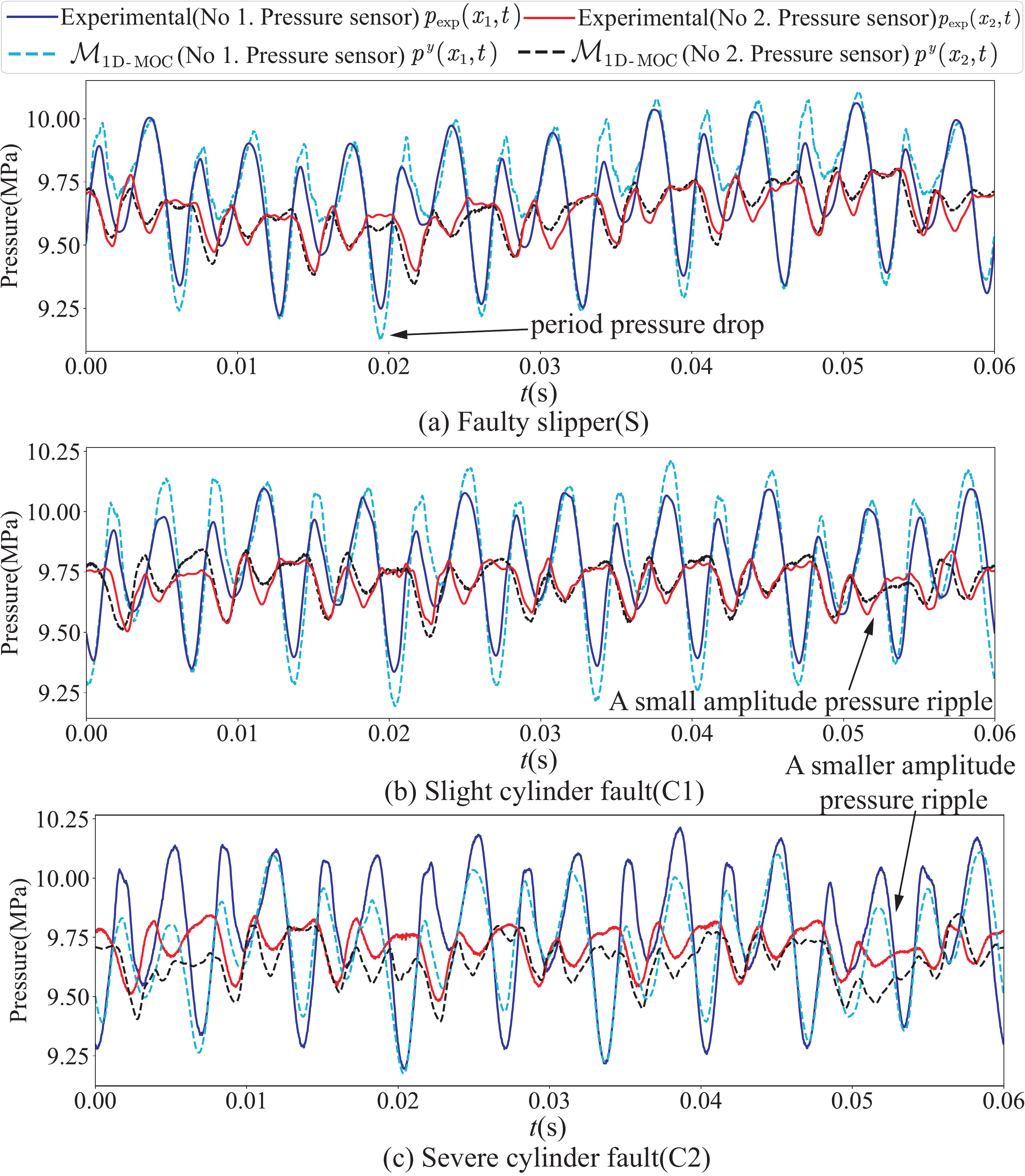}
		\caption{Pressure ripple comparison under faulty conditions: experimental data versus simulation results.}
		\label{fig_pressure_ripple}
	\end{figure}

\begin{figure}[!h]
		\centering	
		\includegraphics[width=1\columnwidth]{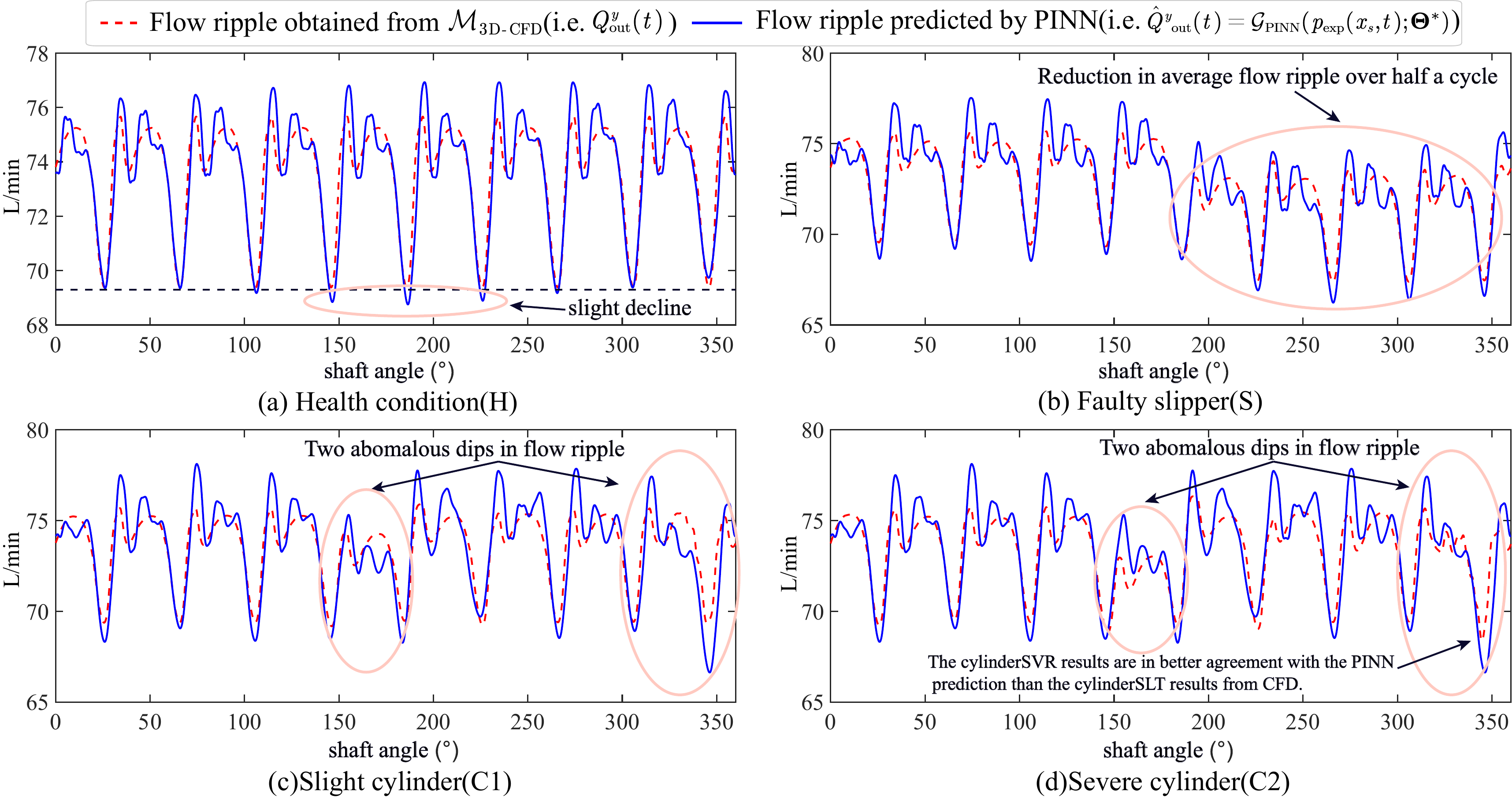}
		\caption{Flow ripple comparison under faulty conditions: experimental data versus simulation results.}
		\label{fig_flow_ripple}
	\end{figure}

	To facilitate synthetic fault dataset construction, four health states were simulated using the calibrated model $\mathcal{M}_{\text{1D-MOC}}$: healthy (H), slipper wear (S), slight cylinder wear (C1), and severe cylinder wear (C2). Faults were introduced by modifying the geometry to incorporate leakage paths, such as scratches on the slipper or cylinder block\cite{dong2023inverse}. Fig.\ref{fig_pressure_ripple} and Fig.\ref{fig_flow_ripple} show a comparison of the simulated and experimental pressure ripple signals and flow ripple signals under fault conditions, respectively. While experimental pressure ripple signals are relatively accessible, acquiring high-frequency flow ripple signals directly poses significant challenges. Therefore, the “experimental” flow ripple data shown in Fig.\ref{fig_flow_ripple} is sourced from our prior work\cite{dong2025innovative}, which utilized a PINN framework predicting outlet flow ripples at the pump outlet.
	
As shown in the Fig.\ref{fig_pressure_ripple} and Fig.\ref{fig_flow_ripple}, both pressure ripple and flow ripple signal exhibit good agreement with experimental signals under healthy condition and faulty slipper condition.  However, discrepancies arise in the case of cylinder faults. In the pressure ripple simulation, the experimental cylinder fault signal corresponds more closely with the slight cylinder fault, whereas in the flow ripple, the PINN prediction aligns better with the severe cylinder fault. When the experimental pressure ripple signal for the cylinder fault is input, the PINN-predicted flow ripple exhibits a more pronounced drop in the second drop, while the 3D CFD simulation of the slight cylinder fault shows minimal secondary decline. Conversely, in terms of pressure ripple, the experimental signal aligns more closely with the slight cylinder fault. This discrepancy may stem from insufficient simulation accuracy of the model for cylinder faults, though the trends in fault feature changes remain consistent. For instance, a smaller pressure ripple subsequence appeared in the pressure ripple signal, and the flow ripple displays two abnormal drops due to the unexpected interconnection of two piston chambers within one cycle. Consequently, in pressure ripple-based diagnostic models, classifying the experimental cylinder fault as a slight cylinder fault is deemed accurate. For flow ripple-based diagnostic models, classifying the experimental cylinder fault signal as a severe cylinder fault is considered correct.

	\begin{figure}[!h]
	\centering
	\subfloat[Health condition\label{fig_4_9_a}]{
		\includegraphics[width=0.32\textwidth]{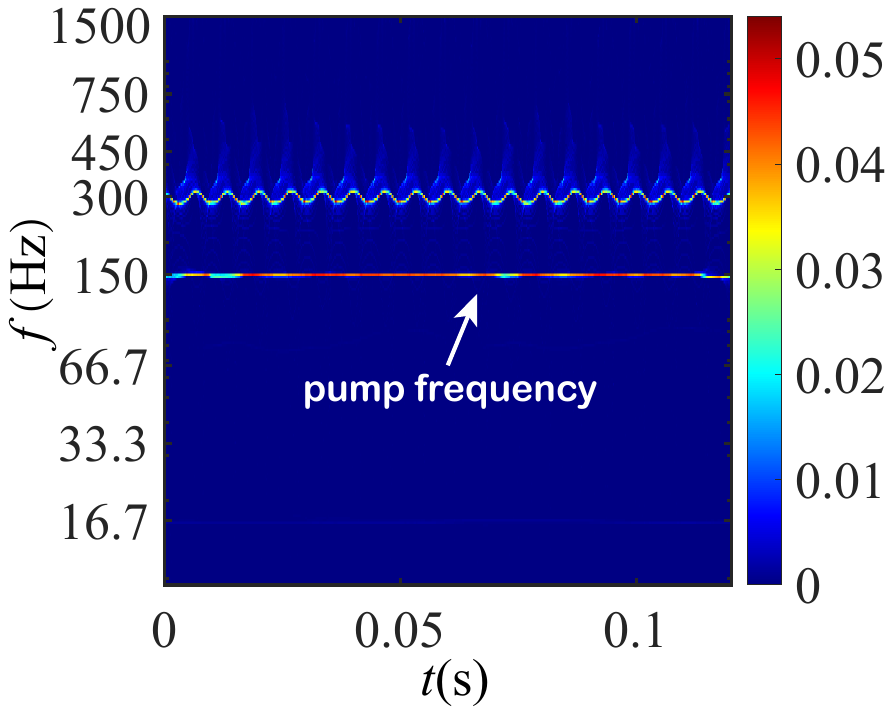}}
	\hfill
	\subfloat[Faulty slipper\label{fig_4_9_b}]{
		\includegraphics[width=0.32\textwidth]{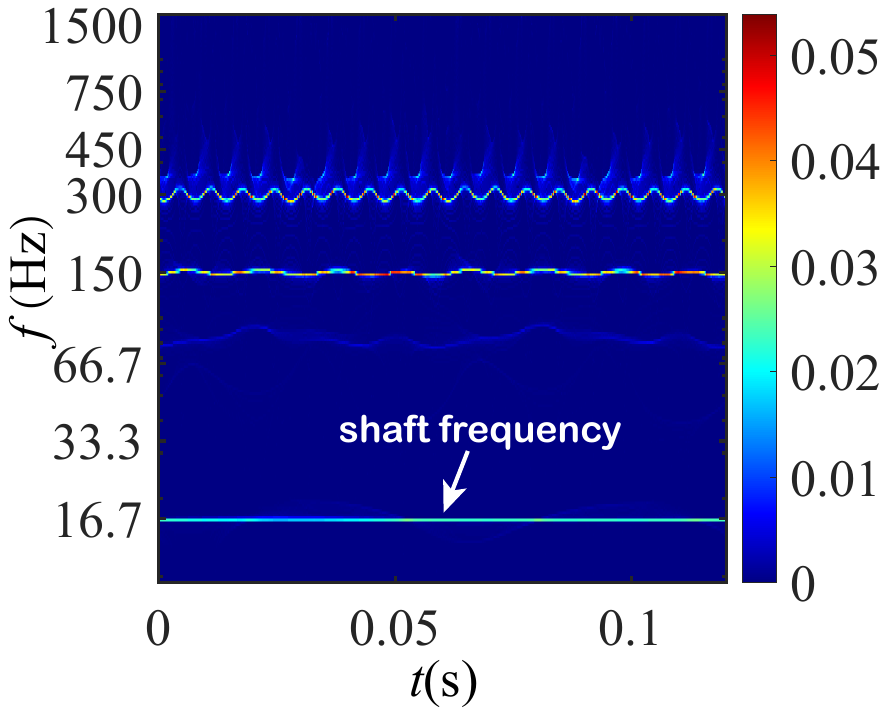}}
	\hfill
	\subfloat[Faulty cylinder\label{fig_4_9_c}]{
		\includegraphics[width=0.32\textwidth]{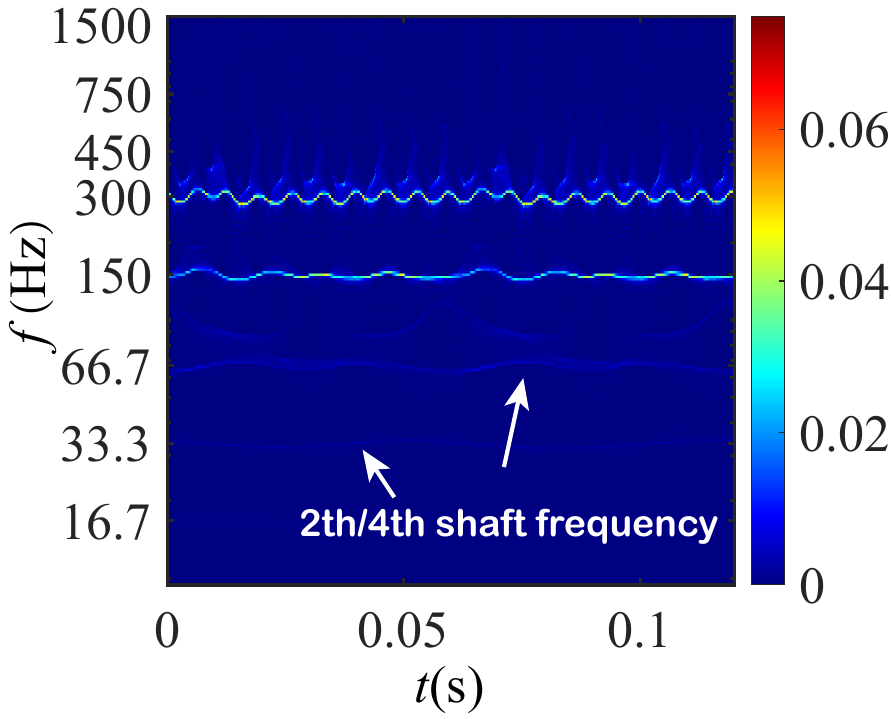}}
	\\
	\caption{The SST spectrograms of experimental pressure ripple $p_{exp}^y(x_1,t)$ }
	\label{fig_sst_exp}
\end{figure}

\begin{figure}[!h]
	\centering
	\begin{minipage}{0.65\textwidth}
		\centering
		\subfloat[Health Condition]{
			\includegraphics[width=0.49\textwidth]{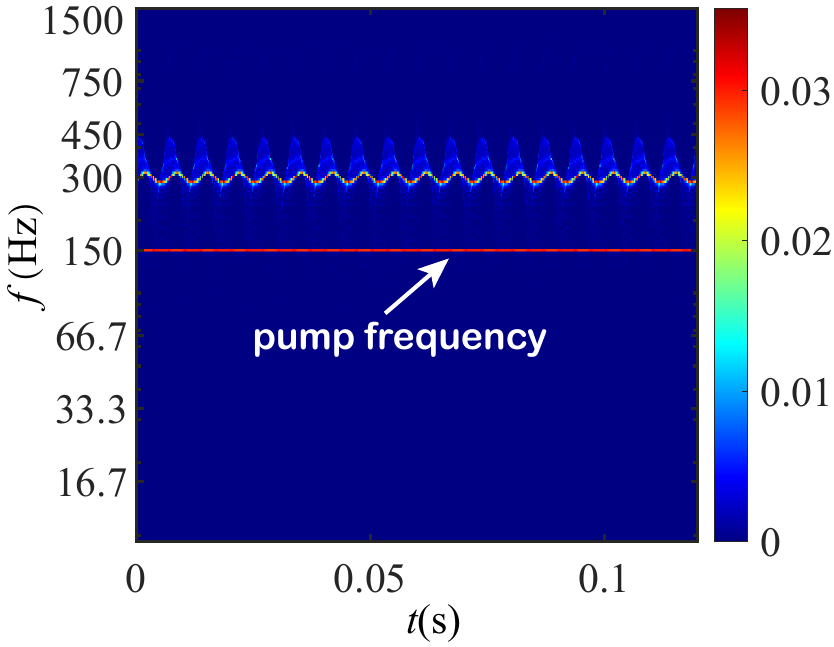}}
		\subfloat[Faulty Slipper Condition]{
			\includegraphics[width=0.49\textwidth]{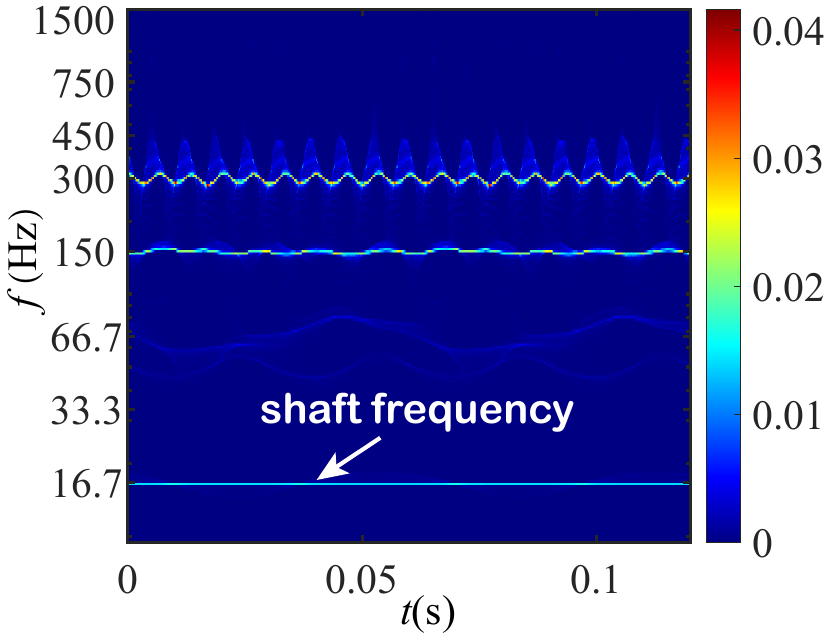}}
		\hspace{0.5em} 
		\subfloat[Slight faulty cylinder Condition]{
			\includegraphics[width=0.49\textwidth]{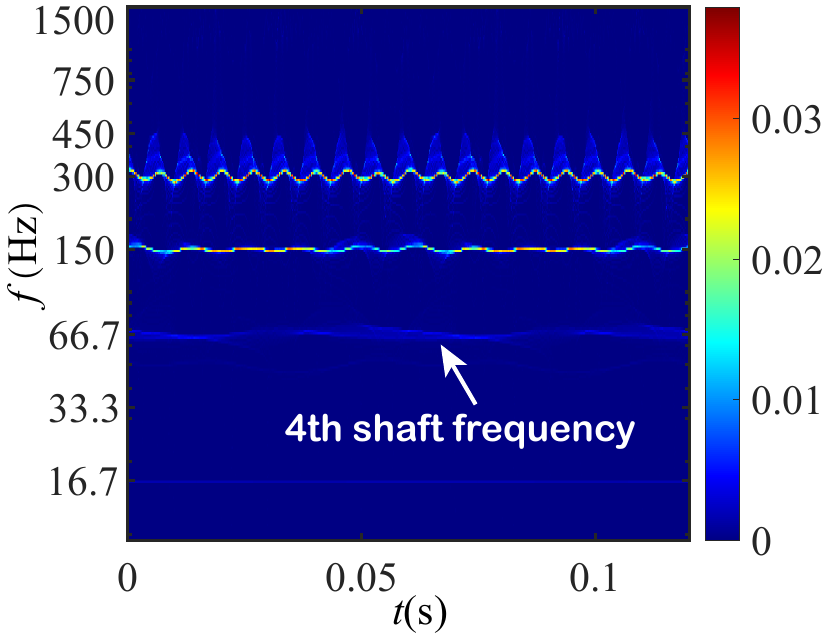}}
		\subfloat[Severe faulty cylinder Condition]{ 
			\includegraphics[width=0.49\textwidth]{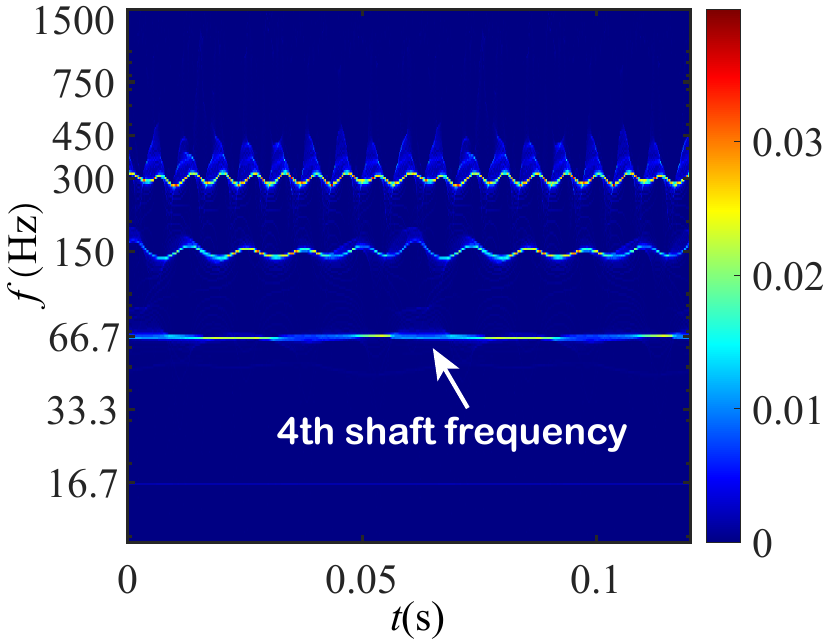}}
	\end{minipage}
	\caption{The SST spectrograms of simulated pressure ripple $p^y(x_1,t)$ from $\mathcal{M}_{\text{1D-MOC}}$ }
	\label{fig_sst_MOC}
\end{figure}

\begin{figure}[!h]
	\centering
	\begin{minipage}{0.65\textwidth}
		\centering
		\subfloat[Health Condition]{
			\includegraphics[width=0.48\textwidth]{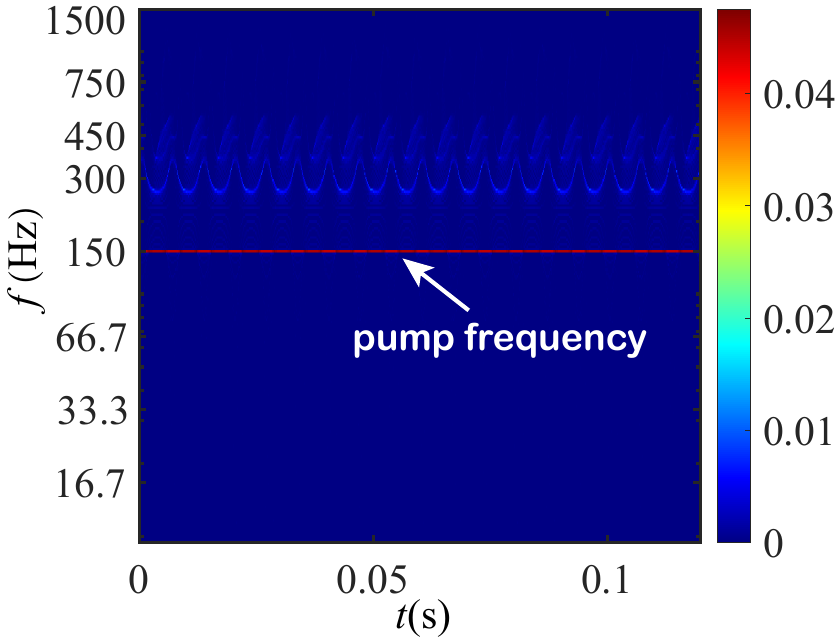}}
		\hfill
		\subfloat[Faulty Slipper Condition]{
			\includegraphics[width=0.48\textwidth]{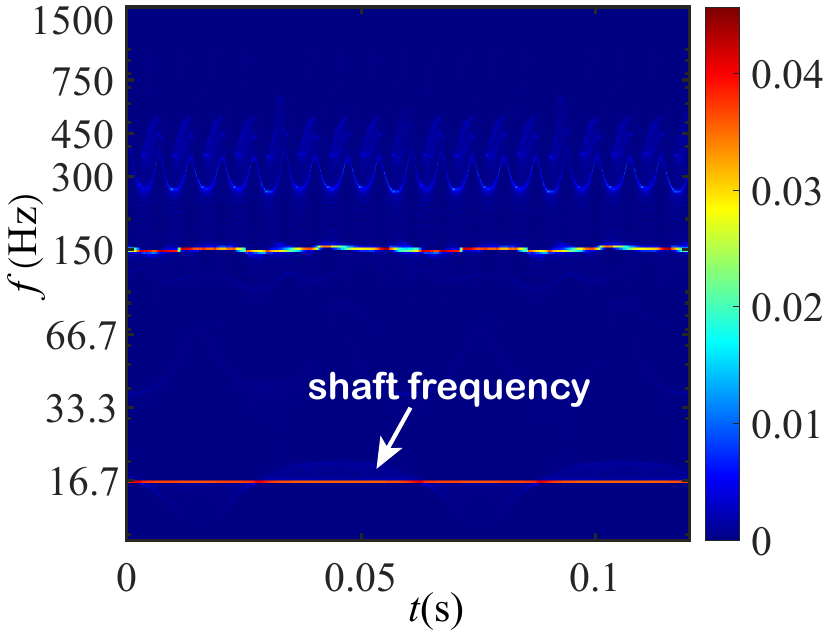}}
		\hfill
		\subfloat[Slight faulty cylinder Condition]{
			\includegraphics[width=0.48\textwidth]{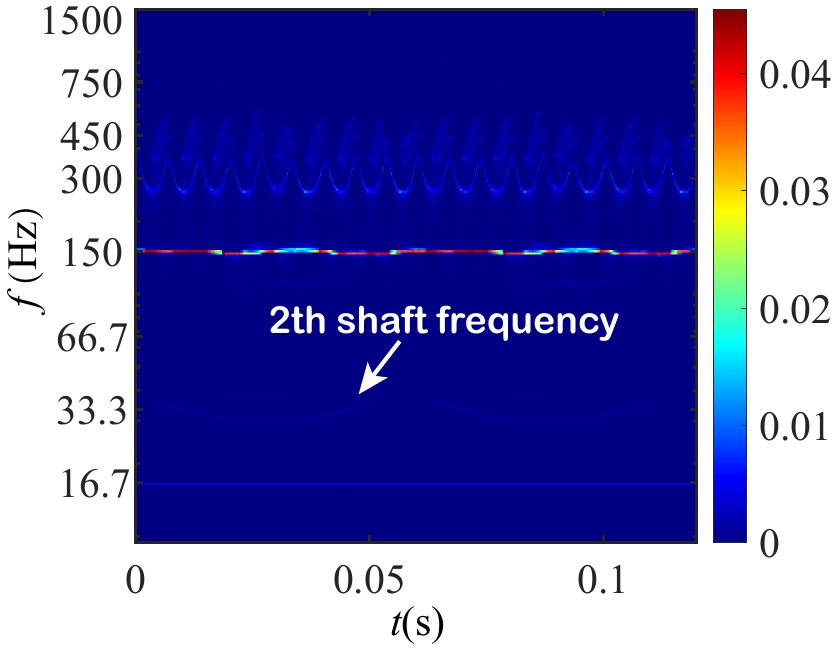}}
		\hfill
		\subfloat[Severe faulty cylinder Condition]{
			\includegraphics[width=0.48\textwidth]{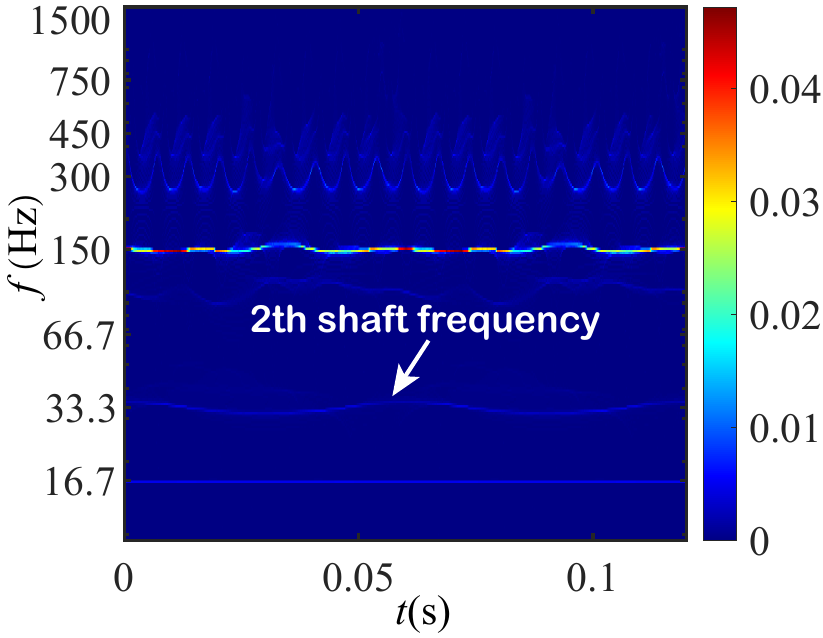}}
	\end{minipage}
	\caption{The SST spectrograms of simulated pressure ripple $p^y(x_1,t)$ from $\mathcal{M}_{\text{3D-CFD}}$ }
    \label{fig_sst_CFD}
\end{figure}

Figs.\ref{fig_sst_exp}, \ref{fig_sst_MOC}, and \ref{fig_sst_CFD} present the SST spectrograms of experimental pressure ripples $p_{\text{exp}}^y(x_1,t)$ and simulated pressure ripples $p^y(x_1,t)$ from the calibrated $\mathcal{M}_{\text{1D-MOC}}$ and uncalibrated $\mathcal{M}_{\text{3D-CFD}}$, respectively. The frequency range is limited to 1500 Hz (10 times the pumping frequency), with logarithmic scaling applied to emphasize shaft and pumping harmonics. The spectrograms are resized to 256×256 for input into 2D CNN models. It is evident that the time-frequency map of the SWT-transformed pressure ripple signals from the uncalibrated model has lower similarity to the experimental signals. In contrast, the calibrated model (Fig.\ref{fig_sst_MOC}) maintains close alignment.  For instance, the energy distribution of uncalibrated model's spectrograms (Fig.\ref{fig_sst_CFD}) at 2× and 3× pumping frequency(i.e. 300Hz, 450Hz) is very weak, whereas the experimental signals have strong energy distributions at two and three times the pumping frequency.  On the other hand, the manifestations of faults on the SWT spectrograms are consistent with the calibrated model and changes in time-series signals. For example, the periodic pressure drop caused by slipper faults results in stable energy distribution at one times the shaft frequency on the time-frequency map. The two internal leakages caused by cylinder faults also show weak energy changes at two and four times the shaft frequency. In the healthy state, the time-frequency map at the pumping frequency of 150 Hz is an almost stable straight line, corresponding to the similarity of pressure ripple subsequences in the healthy state, whereas under fault conditions, fluctuations appear at the 150 Hz pumping frequency, indicating reduced similarity of pressure ripple subsequences.

\begin{table}[!h]
	\caption{Source and construction process of simulation datasets}
	\centering
	\renewcommand{\arraystretch}{1.2}
	\setlength{\tabcolsep}{6pt}
	\begin{tabular}{@{}lp{3.5cm} p{3.5cm} p{6cm}@{}}
		\toprule
		Dataset & Input signal & Data source & Processing procedure \\
		\midrule
		CFD-P & Two-sensor pressure ripples $p^y(x_s, t)$ (e.g.,Fig. \ref{fig_pressure_ripple_H}(a)) & $\mathcal{M}_{\text{3D-CFD}}$  & Collect pressure ripples at sensor locations from $\mathcal{M}_{\text{3D-CFD}}$  under healthy and faulty conditions. Note that due to uncalibrated parameters, simulated signals deviate from experimental ones. \\
		CFD-MOC-P & Two-sensor pressure ripples  $p^y(x_s, t)$ (e.g.,Fig. \ref{fig_pressure_ripple_H}(b) and \ref{fig_pressure_ripple}) &$\mathcal{M}_{\text{1D-MOC}}(Q_{\text{out}}^y(t);\mathbf{\Theta}^*)$  & Collect pressure ripples at sensor locations from $\mathcal{M}_{\text{1D-MOC}}(Q_{\text{out}}^y(t);\mathbf{\Theta}^*)$ under healthy and faulty conditions; Simulated signals closely match experimental ones;\\
		CFD-Q & Flow ripple at pump outlet $Q_{\text{out}}^y(t)$ (e.g., Fig. \ref{fig_flow_ripple}) & $\mathcal{M}_{\text{3D-CFD}}$  & Collect pump outlet flow ripples from $\mathcal{M}_{\text{3D-CFD}}$  under healthy and faulty conditions; During evaluation of unknown samples $p_{\text{exp}}(x_s,t)$, first use the $\mathcal{G}_{\text{PINN}}(p_{\text{exp}}(x_s,t); \mathbf{\Theta}^*)$ with the calibrated 1D model to convert $p_{\text{exp}}(x_s,t)$ into $\hat{Q}_{\text{out}}(t)$, then input into the flow ripple-based diagnostic model. \\
		\bottomrule	 
	\end{tabular}
	\label{table_dataset_description}
\end{table}

Three distinct synthetic datasets were constructed from the simulation models to train neural networks for the four-class (H, S, C1, C2) fault diagnosis task, each corresponding to a different diagnostic input signal as summarized in Table \ref{table_dataset_description}. The CFD-P dataset, comprising pressure ripples from the uncalibrated $\mathcal{M}_{\text{3D-CFD}}$ model, serves as a reference benchmark to evaluate the performance gain achieved by model calibration. In contrast, the CFD-MOC-P dataset contains pressure ripples from the calibrated $\mathcal{M}_{\text{1D-MOC}}$ model, which closely match experimental signals. The CFD-Q dataset provides flow ripples from $\mathcal{M}_{\text{3D-CFD}}$, with its diagnostic pipeline incorporating a PINN-based signal converter for experimental applications. For each of the four health states in every dataset, a single high-fidelity simulated seed signal was augmented to generate 1000 samples. This was achieved through random cyclic shifts and the addition of Gaussian noise with a standard deviation equivalent to 0.15 times the estimated sensor noise, thereby creating a sufficiently large and varied dataset to effectively train deep learning models.
\subsection{Diagnostic performance comparison}

To evaluate the models trained on synthetic data, an experimental test dataset was constructed using pressure ripple signals acquired from a physical test rig. The pump was operated at a constant speed of 1000 r/min (rotational period: 0.06 s), and signals were sampled at 51,200 Hz. For each of the three target health conditions—healthy (H), faulty slipper (S), and faulty cylinder (C)—a continuous 9.6-second recording was obtained. To create inputs for time-domain models, each recording was segmented into non-overlapping windows of one rotational period, with each window containing 3072 data points. This yielded a total of 480 samples (160 per class). For time-frequency analysis, SST spectrograms were generated using a two-rotation window (0.12 s, 6144 points) to improve frequency resolution, producing 240 SST samples in total. 

\begin{table}[!h]
	\caption{CNN1D(341@100) Network Architecture and Parameter Settings}{Architecture and Parameter Configuration of CNN1D (341@100)}
	\renewcommand\arraystretch{1.2}
	\centering
	\begin{tabular}{ccccc}
		\toprule
		Layer & Key Parameters & Input Shape & Output Shape & Activation \\
		\midrule
		Conv1D & Kernel size 341, stride 100, channels 64 & (2, 3072) & (64, 59) & ReLU \\
		MaxPool1d & Kernel size 3, stride 2 & (64, 59) & (64, 30) & - \\
		Conv1D & Kernel size 3, stride 1, channels 64 & (64, 30) & (64, 30) & ReLU \\
		MaxPool1d & Kernel size 3, stride 1 & (64, 30) & (64, 30) & - \\
		GAP & - & (64, 30) & (64) & - \\
		SoftMax & - & (64) & (4) & SoftMax \\
		\bottomrule
	\end{tabular}
	\label{table_6_network_setup}
	\begin{tablenotes}
		\footnotesize
		\item [1] CNN1D(341@100): Indicates a 1D CNN architecture where the first convolutional layer employs a kernel size of 341 and a stride of 100; the CNN1D(150@50) and CNN1D(3@1) architectures follow a similar naming convention.
		\item [2] ResNet18(341@100): Denotes an architecture in which the first convolutional layer uses a kernel size of 341 and stride of 100, while the subsequent framework adopts residual blocks similar to the standard ResNet18 design.
	\end{tablenotes}
\end{table}

For time-series inputs, one-dimensional CNN serves as the primary architecture. The first layer of CNN is generally regarded as the feature extraction layer, crucial for classification results. Three designs with different receptive field sizes were selected for this study. The network architecture of the 1D large-kernel CNN (kernel size 341, stride 100) is shown in Table \ref{table_6_network_setup}; CNN1D(3@1) and CNN1D(150@50) differ only in the first convolutional layer parameters, with the rest identical to CNN1D(341@100). ResNet architectures were chosen to implement deeper networks to explore the impact of network depth on the accuracy of simulation-driven fault diagnosis models. As shown in Table \ref{table_6_network_setup}, CNN1D(341@100) is a 6-layer network design, while ResNet18(341@100) denotes an 18-layer architecture via residual connections, with the first layer consistent with CNN1D(341@100).  Two-dimensional CNN serves as the primary network architecture for time-frequency processing, with the architecture and parameters of CNN2D(3$\times$3@1) shown in Table \ref{table_wdcnn2d_network_setup}. Similar to time-series inputs, three different feature layer parameters and network depths were selected for ablation experiments.

\begin{table}[!h]
	\caption{CNN2D(3$\times$3@1) Network Architecture and Parameter Settings}{Architecture and Parameter Configuration of CNN2D (3$\times$3@1)}
	\renewcommand\arraystretch{1.2}
	\centering
	\begin{tabular}{ccccc}
		\toprule
		Layer & Key Parameters & Input Shape & Output Shape & Activation \\
		\midrule
		Conv2d & Kernel size 3$\times$3, stride 1, channels 64 & (2, 256, 256) & (64, 256, 256) & ReLU \\
		MaxPool2d & Kernel size 2$\times$2, stride 2 & (64, 256, 256) & (64, 128, 128) & - \\
		Conv2d & Kernel size 3$\times$3, stride 1, channels 64 & (64, 128, 128) & (64, 128, 128) & ReLU \\
		MaxPool2d & Kernel size 2$\times$2, stride 2 & (64, 128, 128) & (64, 64, 64) & - \\
		GAP & - & (64, 64, 64) & (64, 1, 1) & - \\
		SoftMax & Input 64, output 4 & (64) & (4) & - \\
		\bottomrule
	\end{tabular}
	\begin{tablenotes}
		\footnotesize
		\item [1] CNN2D(3$\times$3@1): Represents a 2D CNN architecture in which the convolutional layers use a kernel size of 3$\times$3 and a stride of 1. Architectures denoted as \texttt{CNN2D(7$\times$7@2)} and \texttt{CNN2D(30$\times$30@3)} follow the same naming convention.
		\item [2] ResNet18(3$\times$3@1): Denotes a variant where the initial convolutional layer employs a 3$\times$3 kernel with stride 1, while the remaining structure adopts residual blocks consistent with the standard ResNet18. ResNet34(3$\times$3@1) variant follows an analogous pattern.
	\end{tablenotes}
	\label{table_wdcnn2d_network_setup}
\end{table}

	To eliminate the influence of network architecture classification capability, ablation experiments (source domain: experimental dataset) were conducted. Each network architecture was also trained on the experimental dataset and tested on the experimental test set to verify basic classification capability. The training-test split ratio for this ablation experiment is 8:2. For all experiments (source domain: simulation dataset, source domain: experimental dataset) and all network architectures, 10 completely independent dataset generations/splits/training/tests were performed to ensure the correctness and reliability of the results. All performance metrics reported for model evaluation represent the mean ± standard deviation obtained from 10 independent runs with varying random seeds to ensure statistical robustness. Table \ref{table_fault_diganosis_result} presents a comparison of diagnostic accuracies across different architecture and training dataset.
			
	\begin{table}[!h]
		\caption{Diagnostic Accuracy Comparison}
		\renewcommand\arraystretch{1.2}
		\centering
		\setlength{\tabcolsep}{8pt}
		\begin{tabular}{@{}p{2.3cm}p{3.5cm}p{1.5cm}cp{1.4cm}cc@{}}
			\toprule
			\multirow{2}{*}{Input} & \multirow{2}{*}{Architecture} & \multicolumn{3}{c}{Training Dataset: Simulation Dataset} & \multirow{2}{*}{\parbox{3cm}{\centering Training Dataset: \\ Experimental Dataset}} \\
			\cmidrule(lr){3-5}
			& & CFD-P & CFD-MOC-P & CFD-Q & \\
			\midrule
			\multirow{5}{*}{Time Domain}
			& CNN1D (341@100)      & 37.6$\pm$25.6 & \textbf{100$\pm$0.00} & 99.1$\pm$0.99 & 100$\pm$0.00 \\
			& CNN1D (150@50)       & 0.00$\pm$0.00 & 82.7$\pm$14.1        & 64.1$\pm$5.14 & 100$\pm$0.00 \\
			& CNN1D (3@1)          & 28.0$\pm$10.7 & 66.8$\pm$0.19         & 33.3$\pm$0.00 & 100$\pm$0.00 \\
			& ResNet18 (341@100)   & 9.00$\pm$14.2 & 99.4$\pm$1.21        & 89.5$\pm$10.4 & 100$\pm$0.00 \\
			& ResNet34 (341@100)   & 11.8$\pm$19.5 & 78.2$\pm$19.9         & 76.7$\pm$9.07 & 100$\pm$0.00 \\
			\midrule
			\multirow{5}{*}{\makecell{Time-Frequency \\ Domain}}
			& CNN2D (3$\times$3@1)    & 54.0$\pm$19.7 & 99.4$\pm$1.51         & \textbf{100.0$\pm$0.00}& 100$\pm$0.00 \\
			& CNN2D (7$\times$7@2)    & 27.8$\pm$9.47 & 84.0$\pm$13.1         & 85.5$\pm$12.8 & 100$\pm$0.00 \\
			& CNN2D (30$\times$30@3)  & 16.8$\pm$10.4 & 66.5$\pm$0.25         & 82.8$\pm$6.75 & 100$\pm$0.00 \\
			& ResNet18 (3$\times$3@1) & 45.3$\pm$19.2 & \textbf{100.0$\pm$0.00} & 95.0$\pm$8.24 & 100$\pm$0.00 \\
			& ResNet34 (3$\times$3@1) & 36.5$\pm$17.9 & 85.5$\pm$14.3         & 71.5$\pm$8.00 & 100$\pm$0.00 \\
			\bottomrule
		\end{tabular}
		\label{table_fault_diganosis_result}
		\begin{tablenotes}
			\scriptsize
			\item [1] Training Dataset: Simulation Dataset — The AI diagnostic model is trained on the simulation dataset and tested on the experimental dataset (480 samples in total).
			\item [2] Training Dataset: Experimental Dataset — The AI diagnostic model is trained on the experimental training dataset (384 samples) and tested on the experimental test dataset (96 samples), using an 8:2 train-test split.
		\end{tablenotes}
	\end{table}

	First, examining the results from the second paradigm, all network architectures achieved 100\% diagnostic accuracy across 10 runs. This indicates that, under conditions where the source and target domains exhibit minimal distribution differences, all selected architectures are capable of fulfilling the diagnostic task. In scenarios with abundant fault data and consistent distributions, deep learning-based fault diagnosis algorithms can readily achieve high precision classification. However, when distribution discrepancies exist between the source and target domains, markedly different outcomes emerge. Models trained on the uncalibrated CFD-P dataset significantly underperform those trained on the calibrated CFD-MOC-P and CFD-Q datasets. This marked performance gap directly underscores that the fidelity between simulation and experimental signals is paramount for effective simulation-driven diagnosis. For instance, the CNN1D(150@50) architecture yielded 0\% accuracy in all 10 runs. This underscores that the fidelity between simulation and experimental signals is paramount for the effectiveness of simulation-driven diagnosis. Distributional deviations lead the model to misattribute system differences as fault features, thereby degrading diagnostic precision. This also indirectly demonstrates the low usability of pure data-driven methods in practical applications. CFD-P can also be regarded as training on historical fault datasets and then deploying to systems with parameter drifts or new hydraulic systems, where distribution differences necessitate recollecting fault datasets for the new hydraulic system or drifted systems to retrain the model for normal operation. The cost of building new fault datasets is high, leading to the low usability of pure data-driven "black-box" models in practical applications. In contrast, the parameter-calibrated physical model-based method requires only healthy-state data (with acquisition costs far lower than fault data) to adjust the simulation model parameters for deployment. In contrast, on the CFD-MOC-P (calibrated pressure ripple) and CFD-Q (flow ripple) datasets, appropriate model selection yields accuracies exceeding 90\%. For example, the CNN1D(341@100) architecture trained on CFD-MOC-P achieves 100\% accuracy in all 10 runs without requiring any experimental fault data. For CFD-Q, the spectrogram input with CNN2D(3$\times$3@1) architecture  attains 100.0\% accuracy. In CFD-Q's time-domain inputs, misclassifications primarily involve identifying healthy states as slight cylinder faults, attributable to minor downward trends in PINN-predicted healthy flow ripples, whereas the simulation database features highly similar pressure subsequences. Flow ripple-based diagnosis offers high interpretability, allowing hydraulic practitioners to judge and understand the piston pump's state based on the degree and form of flow ripple drops\cite{dong2025innovative}. The slight drop in flow ripple as shown in Fig.\ref{fig_flow_ripple}(a) has minimal impact on the mean volumetric efficiency; mis-alarmed samples can be relabeled as healthy for model fine-tuning, or a threshold for flow ripple drop can be added to avoid alarming such samples. This demonstrates the importance of algorithm interpretability for fault diagnosis and predictive maintenance; interpretable algorithms can assist in labeling monitoring signals and understanding fault conditions and severity. 
	
	The results for time-frequency domain inputs show high similarity to time-domain inputs, for example, both exhibit changes in diagnostic accuracy with adjustments to feature layer (i.e., first convolutional layer) parameters, and network depth also affects accuracy; this part is understood through the next subsection based on Grad-CAM visualization of neural network decision bases.

\subsection{Results Analysis Based on Grad-CAM Visualization}

\begin{figure}[!htbp]
	\centering
	\subfloat[Health condition\label{fig_6_P_Grad_cam_a}]{
		\includegraphics[width=0.9\linewidth]{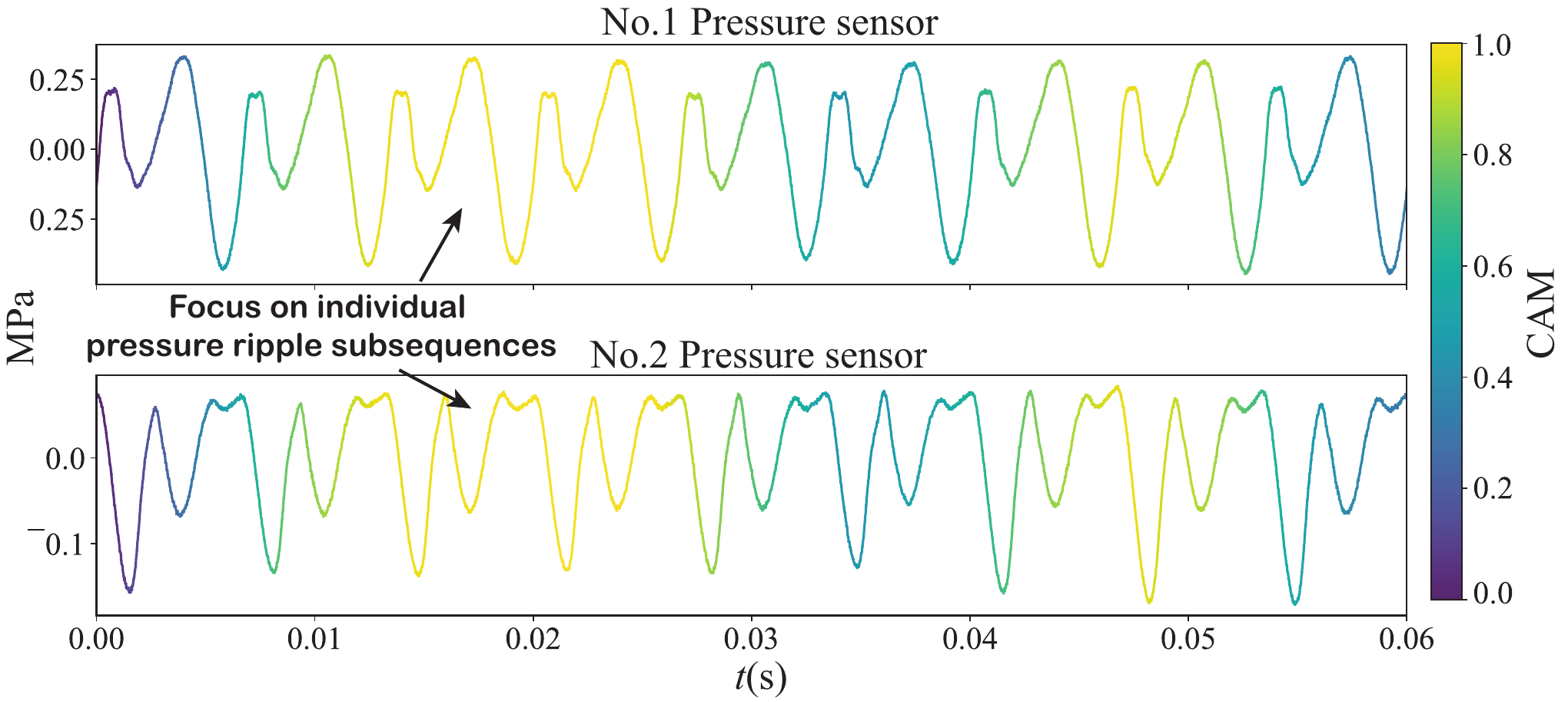}}
	\hfill
	\subfloat[Faulty slipper\label{fig_6_P_Grad_cam_b}]{
		\includegraphics[width=0.9\linewidth]{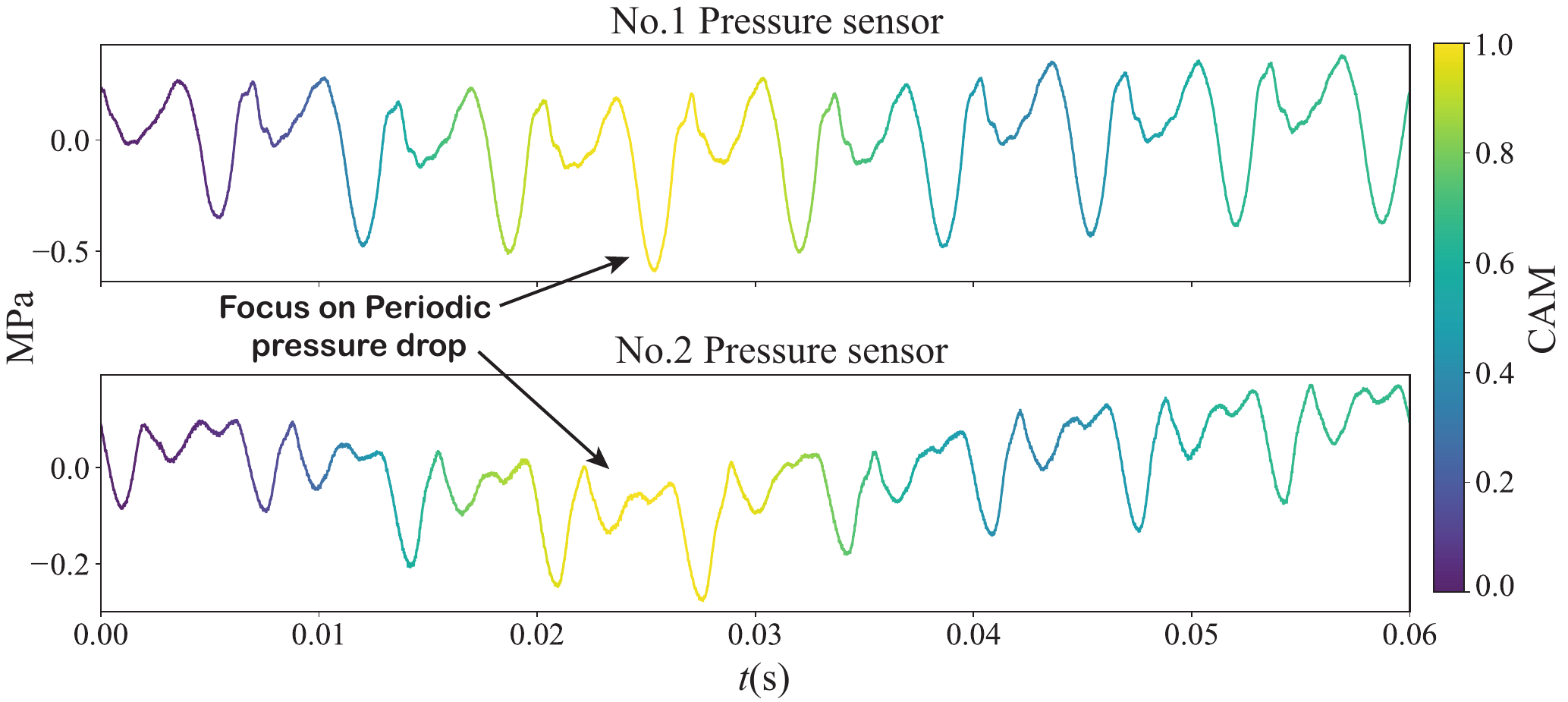}}
	\hfill
	\subfloat[Faulty cylinder\label{fig_6_P_Grad_cam_c}]{
		\includegraphics[width=0.9\linewidth]{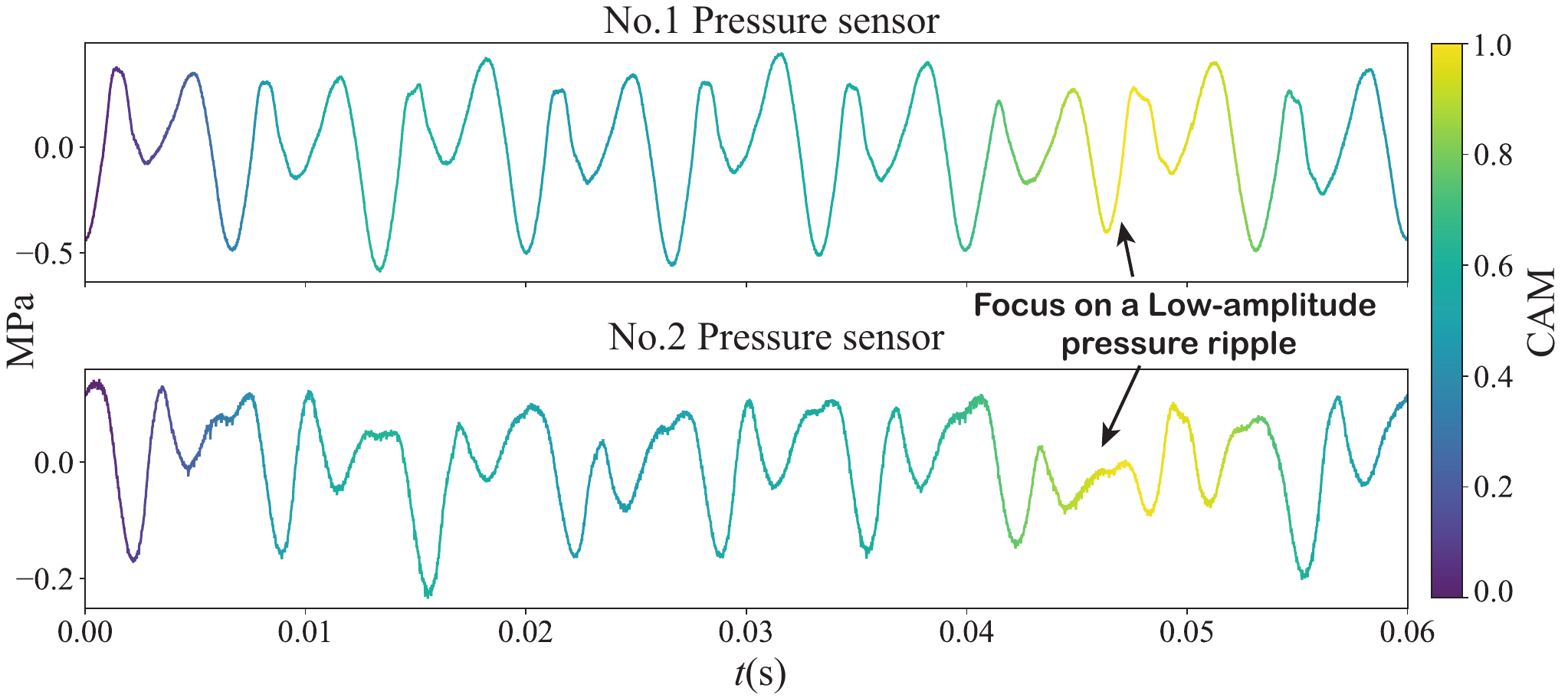}}
	\caption{Grad-CAM-based Visual Explanation for the Decision-Making of CNN1D (341@100) with Pressure Ripple Inputs}
	\label{fig_P_Grad_cam}
\end{figure}

\begin{figure}[!htbp]
	\centering
	\subfloat[Health condition\label{fig_6_P_Grad_cam_2a}]{
		\includegraphics[width=0.9\linewidth]{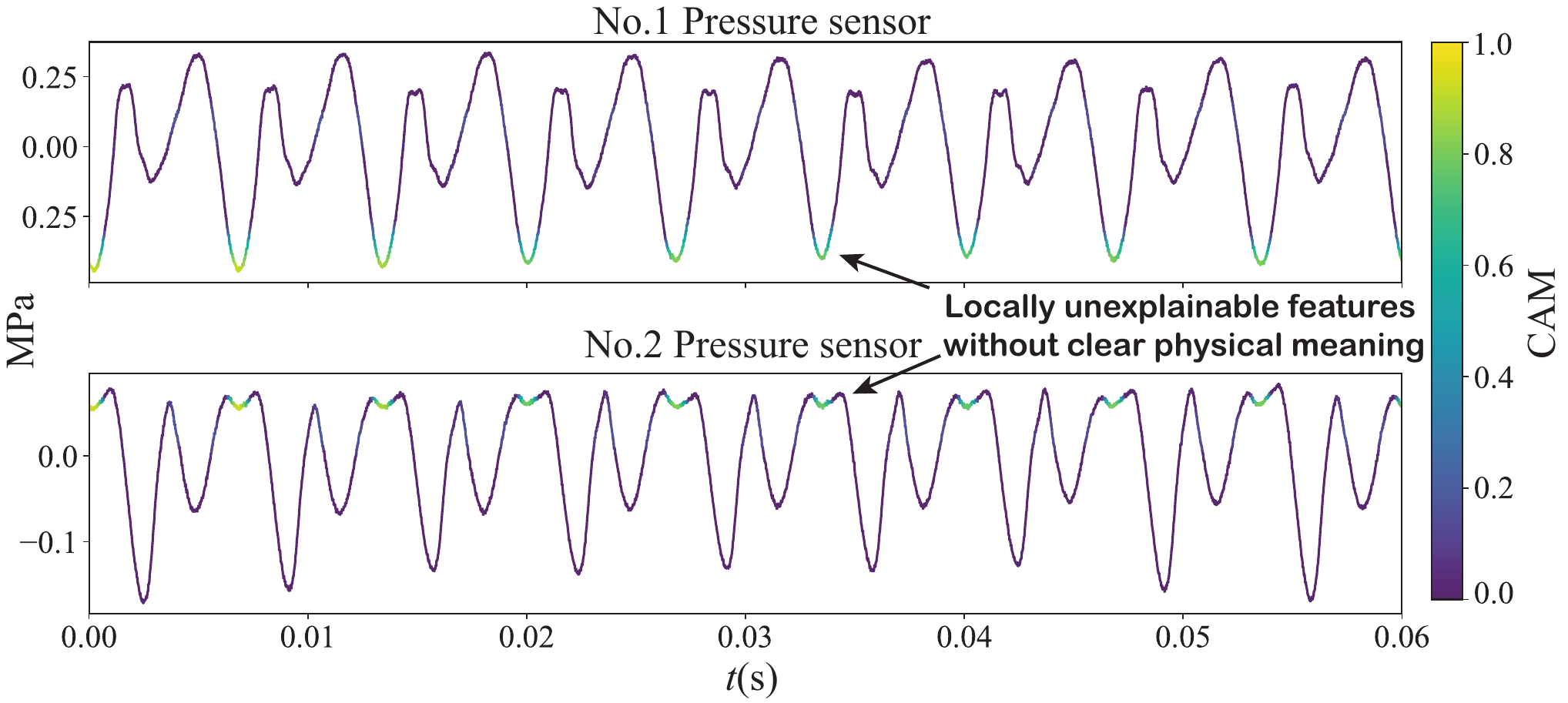}}
	\hfill
	\subfloat[Faulty slipper\label{fig_6_P_Grad_cam_2b}]{
		\includegraphics[width=0.9\linewidth]{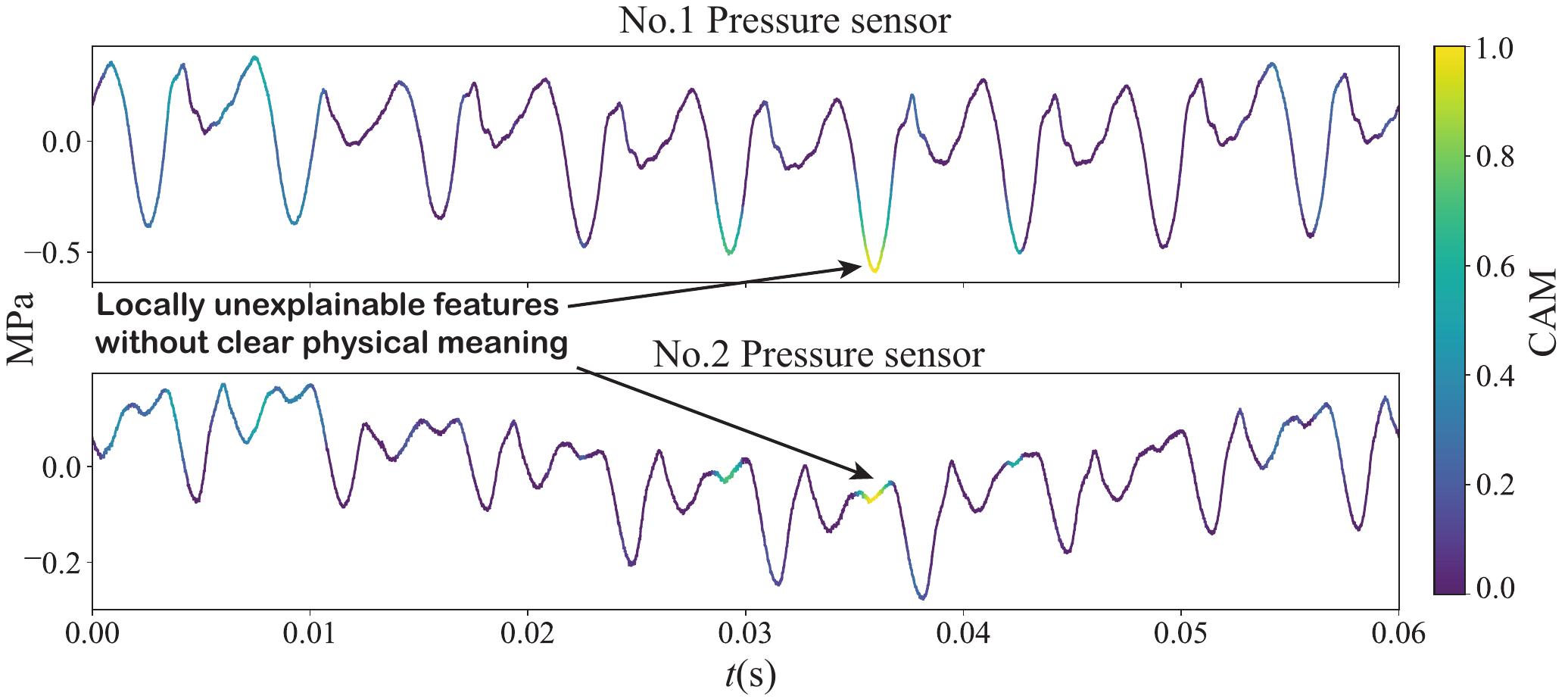}}
	\hfill
	\subfloat[Faulty cylinder\label{fig_6_P_Grad_cam_2c}]{
		\includegraphics[width=0.9\linewidth]{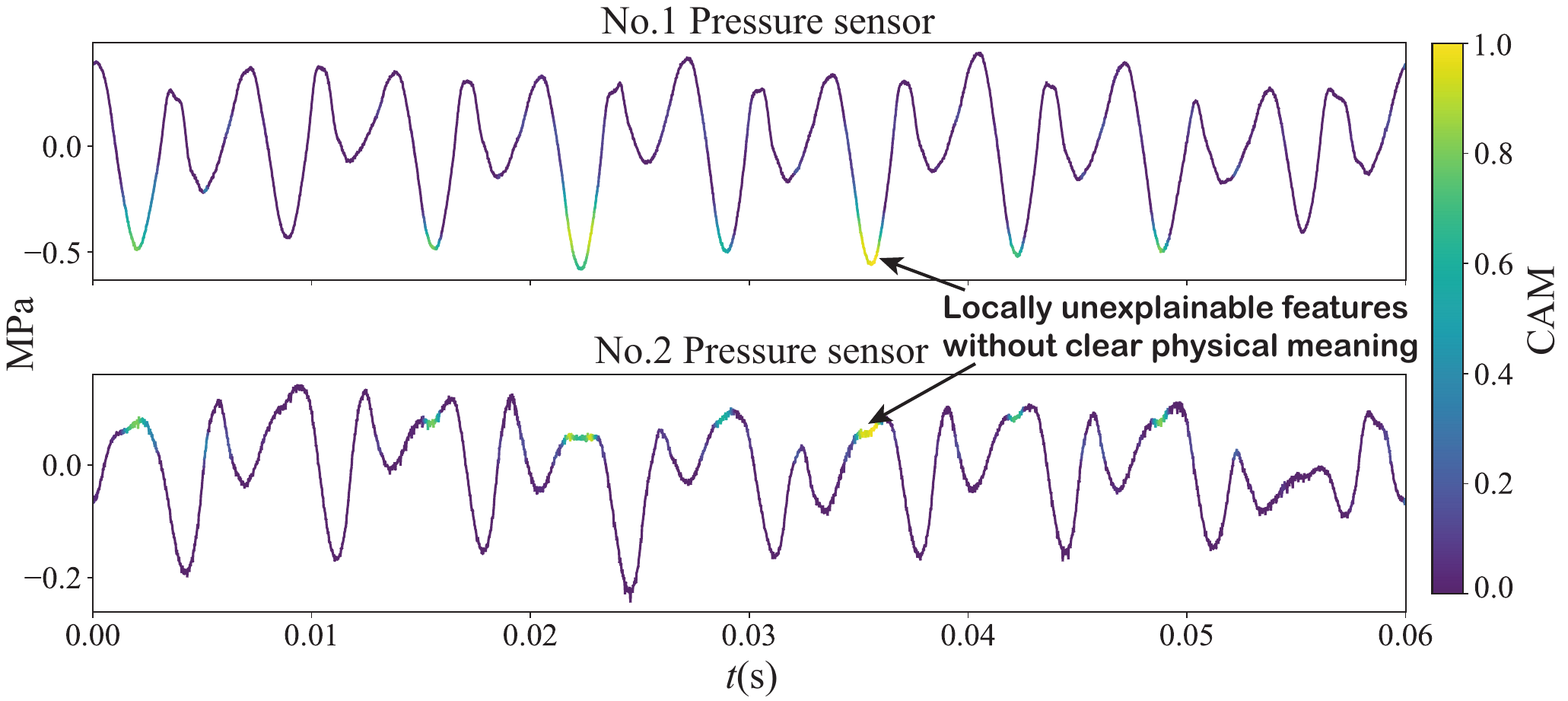}}
	\caption{Grad-CAM-based Visual Explanation for the Decision-Making of CNN1D (3@1) with Pressure Ripple Inputs}
	\label{fig_P_Grad_cam_2}
\end{figure}

Figs \ref{fig_P_Grad_cam} and \ref{fig_P_Grad_cam_2} respectively illustrate the decision processes of different network architectures for time-series inputs via Grad-CAM visualization.

It can be seen that, due to the larger convolutional kernel in CNN1D(341@100), decisions are made based on longer subsequences for fault diagnosis, whereas CNN1D(3@1), with a kernel size of only 3, focuses on local features for decision-making. CNN1D(341@100), trained on the simulation dataset and tested on the experimental dataset, achieves higher accuracy. This is because 341 is actually the length of a single pressure ripple subsequence; as shown in Figs.\ref{fig_P_Grad_cam}(b) and (c), the neural network model essentially judges slipper wear and cylinder wear through pressure ripples dropping to the periodic pressure drop and small-amplitude pressure ripples, fully aligning with the physical laws of the aforementioned fault analysis and modeling\cite{dong2023inverse}. That is, CNN1D(341@100) learns fault features with physical significance. In contrast, the features learned by CNN1D(3@1) are mostly local features without clear physical meaning. However, CNN1D(3@1) trained on the experimental training set and tested on the experimental test set also achieved 100\% diagnostic accuracy in all 10 runs. This is because, although local features lack clear physical meaning, they still enable classification in scenarios with low distribution differences. Nevertheless, when the diagnosis model needs to learn physical knowledge from the simulation model and deploy to actual systems for diagnosis, if the model only learns local features for classification rather than interpretable fault features, and local features lack domain invariance, misjudgments are easily caused. This also explains why CNN1D(341@100) trained on CFD-P has higher accuracy than the other two architectures: even with distribution differences, the model still tends to focus on more physically interpretable features (i.e., features of a single pressure ripple length). This phenomenon also indirectly proves the drawbacks of the non-interpretability of pure data-driven models: in scenarios with low distribution differences (since diagnostic accuracies are all 100\%), it is difficult to distinguish whether the AI model has learned features without physical meaning that only distinguish time series differences, or features with physical meaning that remain accurate when system parameters shift. Therefore, network structure settings closer to a 341 kernel size achieve higher diagnostic accuracy. Unlike previous domain-invariant scenarios of training and testing on experimental data, in simulation-driven fault diagnosis tasks, classification accuracy within a single domain is only one aspect; learning more robust features is a new challenge posed by this scenario.

	\begin{figure}[!h]
	\centering	
	\includegraphics[width=0.85\columnwidth]{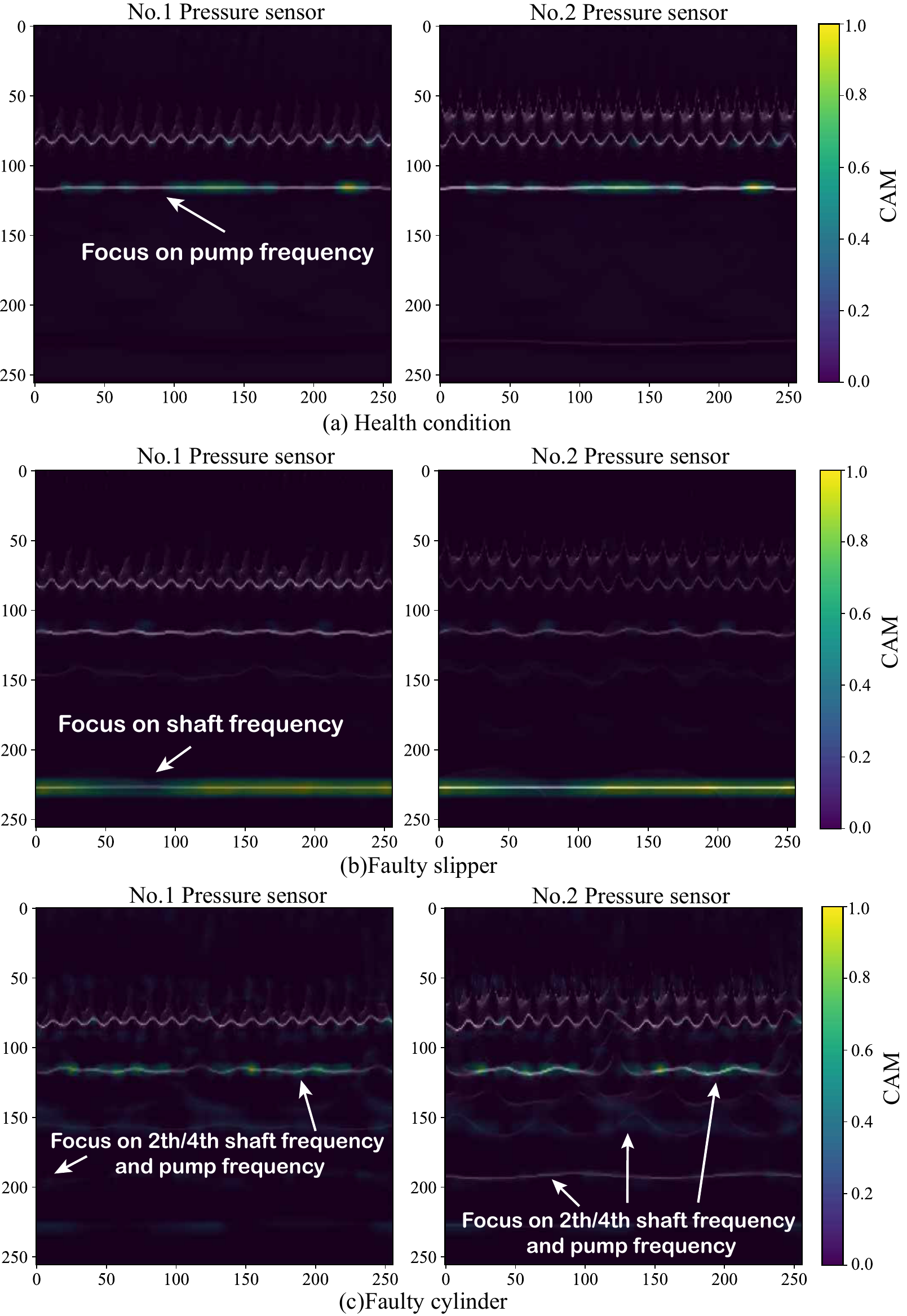}
	\caption{Grad-CAM-based Visual Explanation for the Decision-Making of CNN2D ($3\times3$@1) with Pressure Ripple Inputs}
	\label{fig_P_sst_Grad_cam}
	\end{figure}

	\begin{figure}[!h]
	\centering	
	\includegraphics[width=0.85\columnwidth]{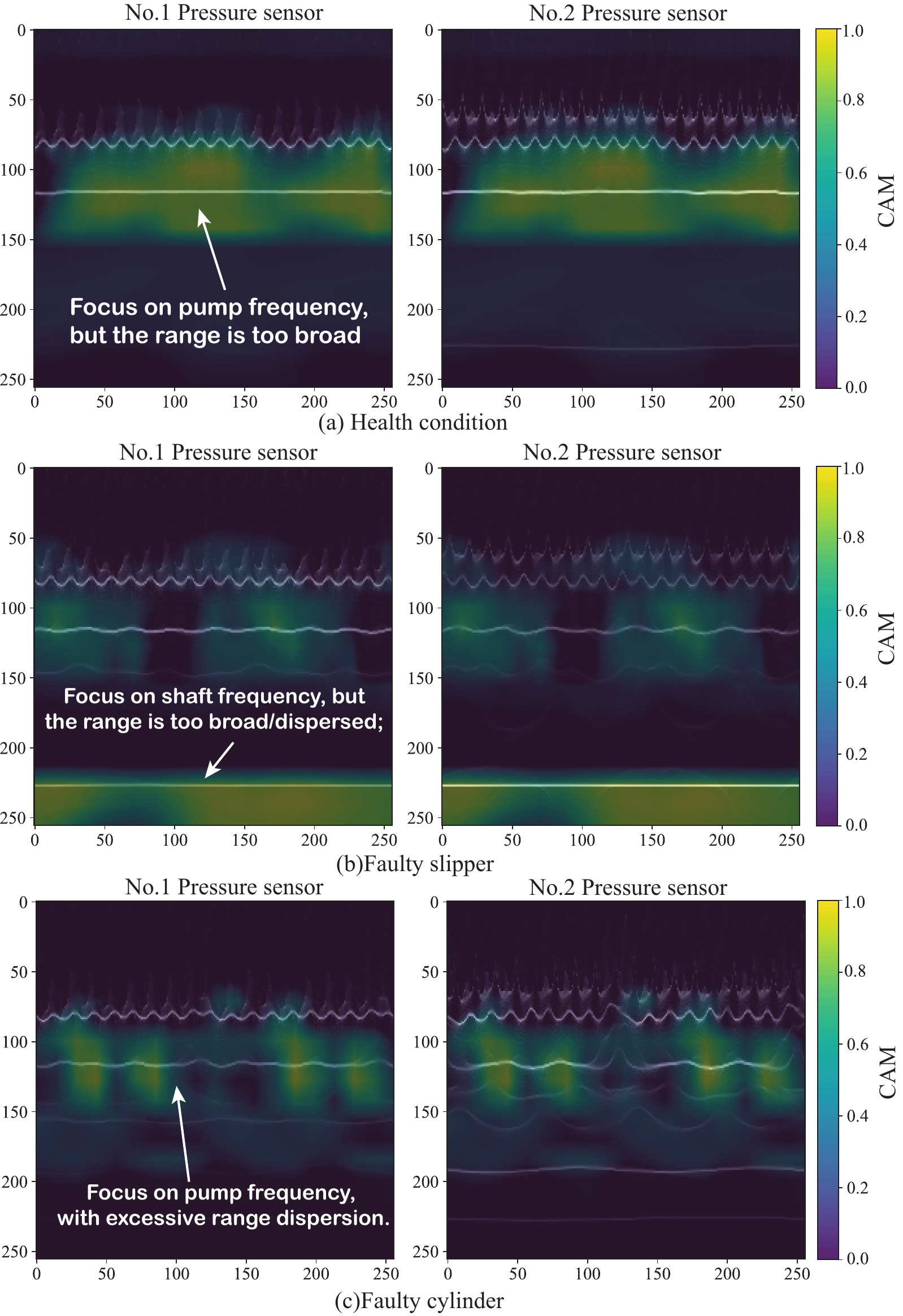}
	\caption{Grad-CAM-based Visual Explanation for the Decision-Making of CNN2D ($30\times30$@3) with Pressure Ripple Inputs}
	\label{fig_P_sst_Grad_cam2}
\end{figure}

In summary, for time-series inputs, larger convolutional kernels facilitate the learning of physically interpretable patterns, evidenced by a consistent trend of increasing diagnostic accuracy with kernel size across all datasets (CFD-P, CFD-MOC-P, CFD-Q). Conversely, experiments with time-frequency domain inputs reveal an inverse relationship: as shown in Table \ref{table_fault_diganosis_result}, enlarging the kernel size in the first convolutional layer leads to broader focus areas and a consequent decline in accuracy. This divergence stems from the fundamental shift in fault representation: in time-frequency maps, localized spectral changes at specific frequencies (e.g., pumping frequency, shaft frequency, and their harmonics) become the physically meaningful signatures, unlike the temporal subsequences salient in raw time-series data. The Grad-CAM visualizations in Figs.\ref{fig_P_sst_Grad_cam} and \ref{fig_P_sst_Grad_cam2} elucidate this mechanism. For time-frequency inputs, smaller kernels stably localize discriminative changes at key frequencies, while larger kernels produce diffuse attention maps, failing to concentrate on these compact, invariant spectral features. In simulation-driven diagnosis, where generalizing to real-world data is paramount, the model must prioritize domain-invariant and interpretable features. Overly broad attention may incorporate non-interpretable or system-specific artifacts, thereby reducing robustness and accuracy—as reflected in the results. This principle also explains the suboptimal performance of deeper architectures like ResNet18 and ResNet34 in this context. Greater network depth and capacity may allow the model to additionally memorize non-generalizable, less interpretable patterns from the simulation data. Consequently, despite their stronger representational power, such models can exhibit reduced robustness when applied to experimental scenarios. Thus, a key challenge in simulation-driven fault diagnosis lies in enhancing neural network depth and capacity while rigorously preserving model interpretability and robustness to domain shifts. 

In the context of hydraulic pump fault diagnosis, as explored in the referenced study,  BO serves as an effective method for hyperparameter tuning in data-rich environments\cite{tang2022intelligent,tang2022synchrosqueezed}. However, this approach is fundamentally data-driven, relying on empirical diagnostic results from labeled experimental fault datasets to evaluate model performance during optimization iterations. Such dependence can introduce biases tied to specific dataset characteristics and necessitates costly fault-injection experiments, rendering it suboptimal for zero-shot or simulation-driven diagnostics where experimental fault data are scarce or unavailable. In contrast, the method adopted in this paper leverages Grad-CAM as a principled mechanism for neural network architecture optimization: it empowers engineers to assess and iteratively refine designs—such as kernel size, depth, and receptive field—based on whether the network's attention aligns with physically interpretable features, thereby achieving superior generalization in simulation-to-reality transfer without requiring experimental fault datasets.

	\begin{figure}[!h]
	\centering	
	\includegraphics[width=0.95\columnwidth]{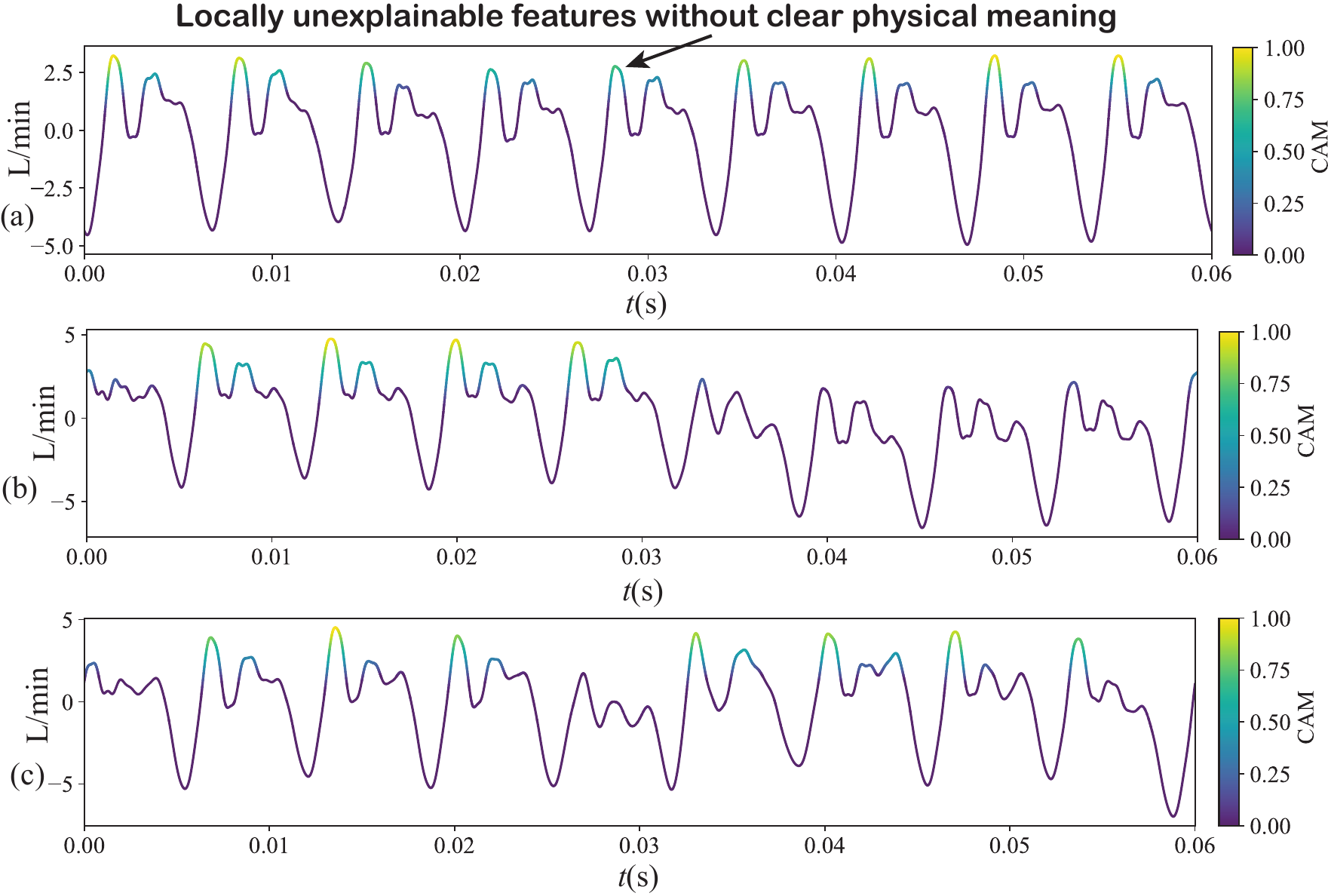}
	\caption{Grad-CAM-based Visual Explanation for the Decision-Making of CNN1D (341@100) with Pump flow ripple Inputs}
	\label{fig_Q_Grad_cam}
    \end{figure}

	\begin{figure}[!h]
	\centering	
	\includegraphics[width=0.95\columnwidth]{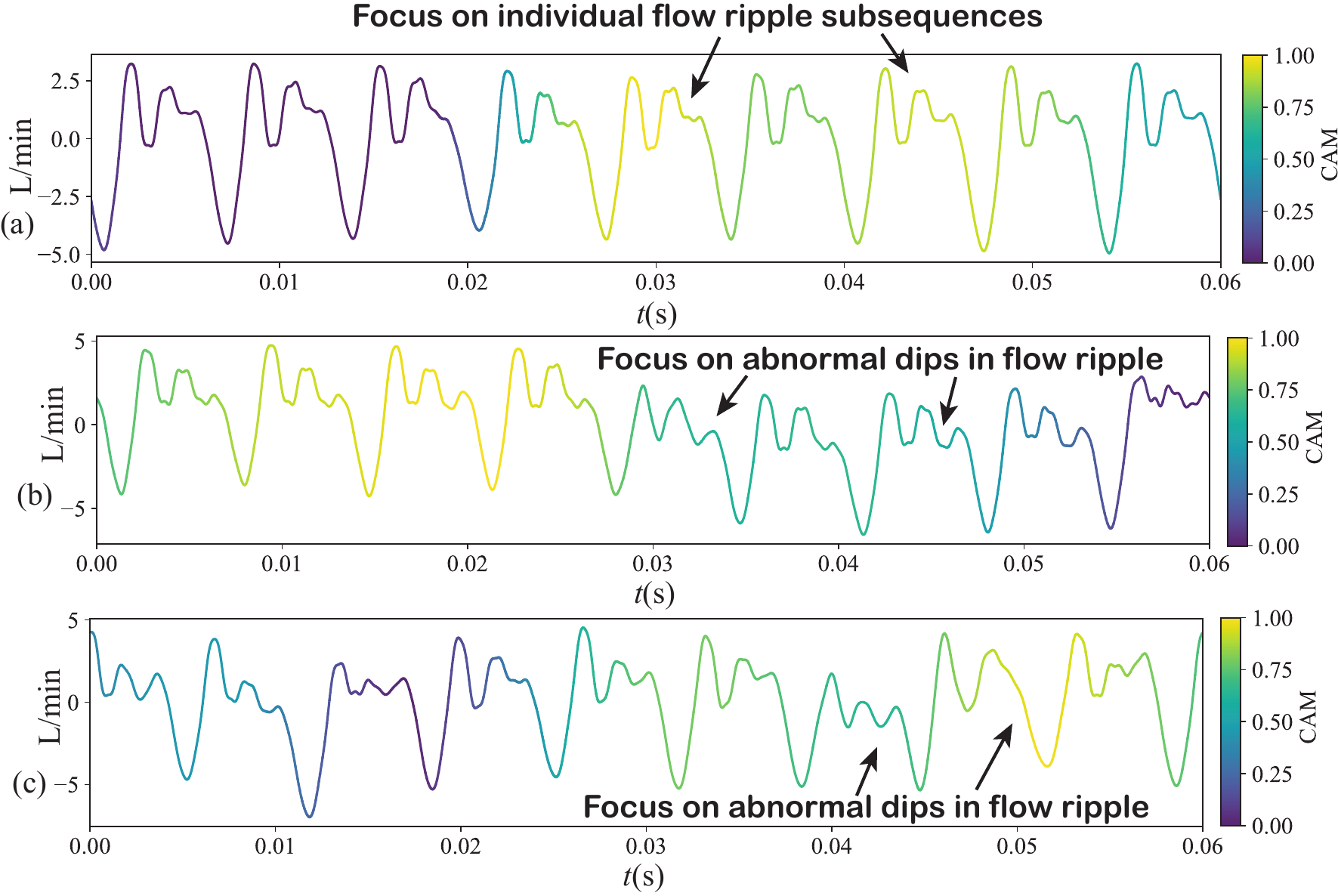}
	\caption{Grad-CAM-based Visual Explanation for the Decision-Making of CNN1D (3@1) with Pump flow ripple Inputs}
	\label{fig_Q_Grad_cam2}
    \end{figure}

	\begin{figure}[!h]
	\centering	
	\includegraphics[width=1\columnwidth]{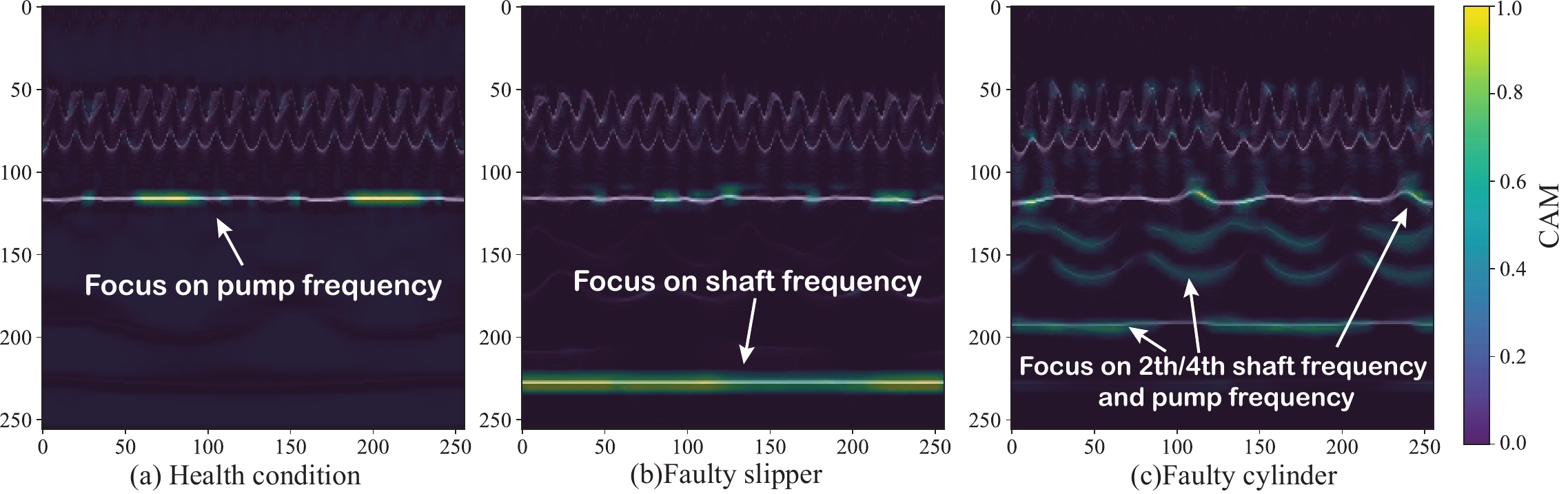}
	\caption{Grad-CAM-based Visual Explanation for the Decision-Making of CNN2D ($3\times3$@1) with Pump flow ripple Inputs}
	\label{fig_Q_sst_Grad_cam}
    \end{figure}

	\begin{figure}[!h]
	\centering	
	\includegraphics[width=1\columnwidth]{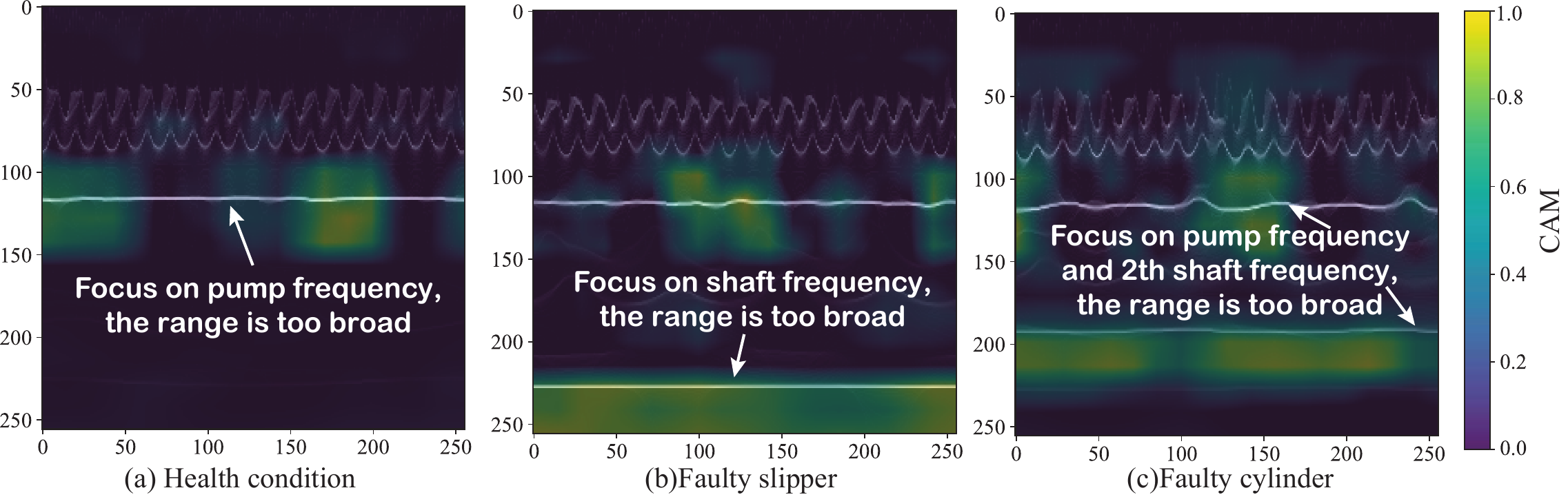}
	\caption{Grad-CAM-based Visual Explanation for the Decision-Making of CNN2D ($30\times30$@3) with Pump flow ripple Inputs}
	\label{fig_Q_sst_Grad_cam2}
    \end{figure}

	The Grad-CAM visualizations based on flow ripple inputs (CFD-Q dataset) reveal a decision-making pattern highly consistent with that observed for pressure ripple inputs. For time-domain flow ripple inputs, as shown in Figs.\ref{fig_Q_Grad_cam} and \ref{fig_Q_Grad_cam2}, the observed pattern aligns completely with the analysis of pressure ripples. The CNN1D (341@100) model with a large convolutional kernel exhibits attention that consistently covers the entire flow-dip waveform within one cycle. This clearly indicates that the model learns to use the “complete flow-drop feature within one cycle” as the physical basis for discriminating between slipper wear and cylinder wear faults—consistent with the underlying flow ripple fault mechanism (e.g., abnormal dips shown in Figs.\ref{fig_flow_ripple}). Such features are highly interpretable and domain-invariant. In contrast, the attention of the small-kernel CNN1D (3@1) model is scattered and focused on local waveform details. Although these details may offer discriminative power within the experimental dataset under matched distributions, they lack clear physical meaning. Consequently, when transferring from the simulation domain (CFD-Q) to the experimental domain, the stability and accuracy of the small-kernel model are significantly lower (see Table \ref{table_fault_diganosis_result}). This reaffirms that, in simulation-driven diagnosis with time domain inputs, guiding the network to learn features that match the physical fault scale (via large kernels in this work) is key to improving model generalizability. In the time–frequency domain, the Grad-CAM visualizations of flow ripple SST spectrograms (Figs.\ref{fig_Q_sst_Grad_cam} and \ref{fig_Q_sst_Grad_cam2}) exhibit the same core trend as observed for pressure ripple time–frequency analysis, and they correspond perfectly with the performance ranking in the CFD-Q column of Table \ref{table_fault_diganosis_result}. The CNN2D model with a small kernel (3×3), which achieves the highest accuracy (100.0\%), demonstrates attention that accurately and steadily localizes changes in key frequency bands caused by faults, such as energy disturbances at the fundamental or harmonic frequencies. In contrast, the CNN2D model with a large kernel (30×30), which shows lower accuracy (82.8\%), produces overly diffuse attention maps that cover excessive spectral regions irrelevant to fault discrimination. This “attention dilution” effect prevents the model from concentrating on the most discriminative, compact spectral features, thereby incorporating irrelevant or system-specific artifacts and directly leading to a decline in diagnostic accuracy.
	
	In summary, whether for pressure or flow ripple inputs, Grad-CAM therefore plays a critical guiding role in neural network design for simulation-driven diagnosis: by revealing attention alignment with physical fault mechanisms, it enables deliberate architectural choices (e.g., large kernels for time-domain, small kernels for time-frequency) that prioritize physically meaningful features, significantly enhancing model robustness and deployability. The success of simulation-driven diagnosis relies on intentionally designing the network architecture (e.g., kernel size) to compel the model to learn these compact, interpretable, and domain-invariant physical features, rather than memorizing non-generalizable details or artifacts potentially present in the simulation data. 
\section{Conclusion}

This study introduces a DT-driven zero-shot fault diagnosis framework for axial piston pumps, leveraging FBN signals to address the challenges of traditional data-driven and model-based methods in data-scarce environments. By calibrating high-fidelity physical models using only healthy-state data and generating synthetic fault signals for training deep learning classifiers, the framework eliminates the need for costly and hazardous experimental fault datasets. The integration of a PINN as a virtual sensor for flow ripple estimation and Grad-CAM for interpretability further enhances the approach, ensuring that model decisions align with underlying physical fault signatures.

Experimental validations on real-world benchmarks demonstrate the framework's efficacy, achieving diagnostic accuracies exceeding 95\% without any experimental fault data. Specifically, the results highlight key insights into network architecture and input signal choices. For time-domain inputs, larger convolutional kernels (e.g., kernel size 341 with stride 100 in CNN1D) outperform smaller ones, yielding 100\% accuracy on calibrated pressure ripple data (CFD-MOC-P) by capturing complete waveform patterns that match the physical fault period, such as periodic pressure drops in slipper wear or dual abnormal dips in cylinder faults. In contrast, smaller kernels (e.g., size 3) focus on local features lacking domain invariance, resulting in lower accuracies (e.g., 66.8\% on CFD-MOC-P). For time-frequency domain inputs, the inverse holds: smaller kernels (e.g., 3×3 in CNN2D) excel, achieving 100\% accuracy on flow ripple data (CFD-Q) by localizing compact spectral changes at key harmonics (e.g., 150 Hz, 300 Hz), while larger kernels (e.g., 30×30) produce diffuse attention maps, reducing accuracy to 82.8\%. Regarding network depth, shallower architectures (e.g., 6-layer CNN1D) consistently surpass deeper ones (e.g., ResNet18 or ResNet34), with the latter showing reduced robustness (e.g., 78.2\% vs. 100\% on CFD-MOC-P time-domain) due to their tendency to memorize non-generalizable patterns from simulation data. Comparisons between pressure and flow ripple inputs demonstrate that, due to the high-fidelity calibrated simulation models exhibiting close alignment with experimental signals, diagnostic accuracies display negligible disparities between the two modalities (e.g., 100\% for pressure ripple using CNN1D(341@100) on CFD-MOC-P versus 100\% for flow ripple using CNN2D(3×3@1) on CFD-Q). Nonetheless, prior research underscores that abnormal declines in flow ripple possess profound physical significance, directly mirroring leakage mechanisms and holding greater promise for fault severity evaluation and enhanced interpretability. These findings, revealed and validated through systematic Grad-CAM analysis, underscore that Grad-CAM is not merely an interpretability tool but a powerful instrument for guiding neural network architecture design. By quantifying attention alignment with physical fault mechanisms, it enables engineers to preferentially select and optimize network parameters (kernel size, receptive field, depth) that consistently focus on physically meaningful features, achieving diagnostic accuracies exceeding 95\% in zero-shot scenarios.

The framework's potential for predictive maintenance in high-stakes applications, such as aerospace and marine systems, is evident, as it bypasses the need for fault-injection experiments. Future research may extend the method to compound faults, incorporate multi-sensor fusion for greater robustness, and investigate real-time DT synchronization to support online monitoring under varying operating conditions.

	\section*{Acknowledgements}  
	This work was supported by the National Key R\&D Program of China under Grant 2023YFB3406705.
	\singlespacing
	\small
	\bibliography{BIB.bib}

\end{document}